\documentclass[ijoc,nonblindrev,12pt]{informs3} 

\OneAndAHalfSpacedXII 

\usepackage{ booktabs, makecell, multirow, multirow, subfig, float}
\usepackage[colorlinks, linkcolor=red, anchorcolor=blue, citecolor=blue]{hyperref}
\usepackage{import}
\usepackage{bbm}
\usepackage{color, xcolor}

\usepackage{graphicx}
\usepackage{amsmath}
\usepackage{amsfonts}

\usepackage{bm,todonotes}
\usepackage{comment}

\newcommand{\cO}{\mathcal{O}}

\newcommand{\EE}{\mathbb{E}}
\newcommand{\RR}{\mathbb{R}}

\newcommand{\cN}{\mathcal{N}}
\newcommand{\cX}{\mathcal{X}}

\newcommand{\hL}{\widehat{L}}

\usepackage{bbm}

\newcommand{\cS}{\mathcal{S}}

\newcommand{\cb}{b}

\newcommand{\norm}[1]{\left\|#1\right\|}
\newcommand{\dotprod}[1]{\left\langle #1\right\rangle}
\def\tr{\mathrm{tr}}

\newcommand{\algname}{\texttt{unknow algorithm to be defined}}

\usepackage{amsmath}

\usepackage{multirow}
\usepackage{natbib}
\usepackage{algorithm}
\usepackage{algorithmic}

 \bibpunct[, ]{(}{)}{,}{a}{}{,}%
 \def\bibfont{\small}%

\newcommand{\xiangyu}[1]{\textbf{\textcolor{red}{[Xiangyu: #1]}}}

\TheoremsNumberedThrough     

\EquationsNumberedThrough    

\begin{document}

\TITLE{An Enhanced Zeroth-Order Stochastic Frank-Wolfe Framework for Constrained Finite-Sum Optimization}

\ARTICLEAUTHORS{

\AUTHOR{Haishan Ye}
\AFF{
	School of Management, Xi'an Jiaotong University, Xi'an, China, 710049 \\ \EMAIL{yehaishan@xjtu.edu.cn}\URL{}}

 \AUTHOR{Yinghui Hang}
\AFF{School of Management, Xi'an Jiaotong University, Xi'an, China, 710049 \\ \EMAIL{yinghui.huang@xjtu.edu.cn}\URL{}}

\AUTHOR{Hao Di}
\AFF{School of Management, Xi'an Jiaotong University, Xi'an, China, 710049 \\ \EMAIL{dihao0836@stu.xjtu.edu.cn}\URL{}}

\AUTHOR{Xiangyu Chang}
\AFF{
		School of Management, Xi'an Jiaotong University, Xi'an, China, 710049\\ \EMAIL{xiangyuchang@xjtu.edu.cn}}
}

\ABSTRACT{%
We propose an enhanced zeroth-order stochastic Frank-Wolfe framework to address constrained finite-sum optimization problems, 
a structure prevalent in large-scale machine-learning applications. 
Our method introduces a novel double variance reduction framework that effectively reduces the gradient approximation variance induced by zeroth-order oracles 
and the stochastic sampling variance from finite-sum objectives. 
By leveraging this framework, our algorithm achieves significant improvements in query efficiency, 
making it particularly well-suited for high-dimensional optimization tasks. 
Specifically, for convex objectives, the algorithm achieves a query complexity of \( \mathcal{O}\left(d\sqrt{n}/\epsilon\right) \) to find an \( \epsilon \)-suboptimal solution, where \( d \) is the dimensionality and \( n \) is the number of functions in the finite-sum objective.
For non-convex objectives, it achieves a query complexity of \( \mathcal{O}\left(d^{3/2}\sqrt{n}/\epsilon^2\right) \) 
without requiring the computation of \( d \) partial derivatives at each iteration. 
These complexities are the best known among zeroth-order stochastic Frank-Wolfe algorithms that avoid explicit gradient calculations. 
Empirical experiments on convex and non-convex machine learning tasks, including sparse logistic regression, robust classification, and adversarial attacks on deep networks, 
validate the computational efficiency and scalability of our approach. 
Our algorithm demonstrates superior performance in both convergence rate and query complexity compared to existing methods.

}


\HISTORY{\today}

\maketitle

%

\section{Introduction}\label{sec:intro}


This paper focuses on the following constrained finite-sum optimization problem:
\begin{equation}\label{eq:prob}
	\min_{x \in \cX} f(x): = \frac{1}{n} \sum_{i=1}^{n} f_i(x),
\end{equation} 
where $f(x):\RR^d \to \RR$ is a smooth function, and $\cX \subset\RR^d$ is the restricted domain. 
This generic form captures numerous modern machine learning (ML) models, ranging from linear models with sparse constraints~\citep{negiar2020stochastic} (e.g., LASSO~\citep{tibshirani1996regression}) to the adversarial attack on deep networks~\citep{chen2020frank}.

To solve the constrained finite-sum optimization problem of~\eqref{eq:prob}, projected (stochastic) gradient descent is a crucial approach \citep{bubeck2015convex}.  
However, projected (stochastic) gradient descent requires a projection onto the constrained set $\cX$, which can be computationally intensive.  
For instance, projection onto the set of all bounded nuclear norm matrices is particularly costly to compute \citep{hazan2016introduction}.  
The Frank-Wolfe algorithm (i.e., conditional gradient) \citep{frank1956algorithm} offers another practical approach to solving problem~\eqref{eq:prob}.  
Unlike projection-based methods, the Frank-Wolfe algorithm only requires solving a linear minimization subproblem in each iteration.  
Furthermore, the linear minimization subproblem in numerous real-world applications is often computationally more efficient than the projection onto $\cX$ \citep{jaggi13}.
Because of the projection-free property and the ability to handle structured constraints, Frank-Wolfe algorithms have been widely studied and applied in various areas, such as robust linear programs \citep{borrero2021modeling,bertsimas2004robust} and matrix completion \citep{allen2017linear}. 
Given the extensive applications of Frank-Wolfe, a recent open-source implementation of several popular Frank-Wolfe variants in the Julia language has been published to assist researchers in developing new approaches \citep{besanccon2022frankwolfe}.

Recently, a specific class of problems involving constrained finite-sum optimization of ~\eqref{eq:prob} has garnered significant attention. 
These problems are characterized by their ``black-box'' nature, in which either the explicit form of the objective function is unknown or, despite having access to the objective function, computational limitations prevent the calculation of gradient information. 
Such challenges are often encountered in real-world scenarios where obtaining gradients is infeasible due to high computational costs (e.g., fine-tuning large language models with limited memory resources~\citep{malladi2023fine}) or restrictions on accessing the analytical form of the function (e.g., black-box adversarial attack~\citep{ilyas2018black}). 

The zeroth-order technique, also known as the derivative-free technique, is an essential tool for addressing scenarios where the analytical form of the objective function is unavailable or the gradient evaluation is computationally prohibitive.
Therefore, the zeroth-order technique has been effectively applied to a variety of ML problems with constrained finite-sum setting, including memory-efficient fine-tuning of large language models \citep{malladi2023fine}, black-box adversarial attacks \citep{ilyas2018black}, reinforcement learning \citep{choromanski2018structured}, and optimization with bandit feedback \citep{bubeck2012regret}.
Moreover, \citet{balasubramanian2018zeroth} propose a zeroth-order stochastic Frank-Wolfe algorithm to improve computational efficiency, which was further enhanced by \citet{sahu2019towards}. \citet{huang2020accelerated} introduce variance reduction techniques to reduce computational costs further, demonstrating the ongoing evolution and refinement of zeroth-order Frank-Wolfe methods in addressing complex optimization.

Despite the extensive research on both the Frank-Wolfe algorithm and zeroth-order optimization, the specific area of zeroth-order Frank-Wolfe algorithms has not been deeply explored, especially in the context of constrained finite-sum optimization problems (as formulated in problem~\eqref{eq:prob}), primarily due to the following challenges.
First, necessitating the use of zeroth-order gradient estimators involves the inherent variance.
Unlike unconstrained settings, where gradients at the optimal point may vanish, 
the constrained nature of the problem ensures that the gradient at the optimal solution remains nonzero \citep{hanzely2018sega,hanzely2020variance}. 
This property, coupled with the inherent variance introduced by zeroth-order gradient estimation, 
leads to a persistent variance that does not diminish over iterations. 
This variance severely impacts the theoretical convergence rate and query complexity of the algorithm, 
making it challenging to achieve efficient optimization \citep{hanzely2018sega,hanzely2020variance}.
Second, large-scale ML applications, which often involve high-dimensional data (\( d \)) and a large number of samples (\( n \)), 
introduce additional difficulties. 
On the one hand, the query complexity and convergence properties of existing zeroth-order Frank-Wolfe algorithms are heavily dependent on \( d \), 
making them computationally prohibitive for high-dimensional problems \citep{sahu2019towards}.
On the other hand, to handle the large sample size \( n \), stochastic optimization techniques such as stochastic gradient descent are commonly employed 
to reduce computational cost \citep{bottou2010large}. 
However, stochastic methods introduce another layer of variance due to random sampling, compounding the optimization challenges. 
To reduce the negative effect of gradient variance induced by zeroth-order oracles, \citet{balasubramanian2018zeroth} try to construct a high precision gradient estimation which takes $\cO(dT)$ function queries for each iteration and achieve a query complexity with lower dependency on the dimension. 
However, due to $\cO(dT)$ function queries for each iteration, the algorithm of \citet{balasubramanian2018zeroth} is not suitable for high-dimension problems.
At the same time, \citet{balasubramanian2018zeroth} does \emph{not} consider the variance of sampling functions, which leads this algorithm to still suffer from a low convergence rate.
\citet{huang2020accelerated} consider both variances and achieve a low total query complexity.
Unfortunately, \citet{huang2020accelerated} still require to compute a high-precision gradient estimation whose construction requires at least $\cO(nd)$ in some iterations. This makes the algorithm of \citet{huang2020accelerated} not applicable for high-dimension problems.

Furthermore, existing research has primarily focused on non-convex objective functions (see \citet{balasubramanian2018zeroth,chen2020frank,sahu2019towards,huang2020accelerated}), leaving a gap in the study of zeroth-order Frank-Wolfe methods explicitly tailored for convex finite-sum problems with constraints. 
Constrained convex optimization is particularly important in practical applications where convergence guarantees and computational efficiency are critical, such as in ML models with structured constraints. 
The absence of a computationally efficient zeroth-order Frank-Wolfe algorithm for convex finite-sum problems thus represents a notable limitation in existing research, one that needs to be addressed to make zeroth-order methods more broadly applicable.
Therefore, the development of an efficient zeroth-order Frank-Wolfe algorithm that can handle the high-dimensional nature of the constrained (convex or non-convex) finite-sum optimization is a compelling direction. 

To address the aforementioned gaps, we develop an enhanced zeroth-order stochastic Frank-Wolfe framework tailored for constrained finite-sum optimization problems, as formulated in \eqref{eq:prob}. 
First, to address the variance introduced by zeroth-order gradient estimation, we design a refined gradient estimator that reduces the persistent variance caused by the constraints. 
This estimator leverages variance reduction techniques to ensure that the variance decreases over iterations without requiring high-precision gradient approximations. 
Second, to mitigate the variance caused by stochastic sampling in finite-sum optimization, we integrate a batch version of refined PAGE method~\citep{li2021page} to balance computational efficiency and variance control. 
The combination of zeroth-order variance reduction and stochastic sampling variance reduction ensures that the algorithm achieves stable convergence in both high-dimensional and large-sample scenarios encountered in ML tasks.
\vspace{-5mm}
\subsection{Related Work}

    \begin{table}[h!]
    \caption{A comparison of query complexity for different algorithms}
    \label{tb:aa}
    \centering
    \scriptsize 
    \begin{tabular}{|c|c|c|c|c|c|c|}
        \hline
        \multirow{2}{*}{Method} 
        & \multicolumn{3}{c|}{Non-convex} 
        & \multicolumn{3}{c|}{Convex} 
        \\ \cline{2-7}
        & Query
        & Max
        & Avg. 
        & Query  
        & Max  
        & Avg. 
        \\ \hline
        \citet{sahu2019towards} 
        & $\cO\left({d^{4/3}}/{\varepsilon^4}\right)$ 
        & $\cO(1)$ 
        & $\cO(1)$ 
        & None 
        & None 
        & None 
        \\ \hline
        \citet{balasubramanian2018zeroth} 
        & $\cO\left({d}/{\varepsilon^4}\right)$ 
        & $\cO(dT)$ 
        & $\cO(dT)$ 
        & $\cO\left({d}/{\varepsilon^3}\right)$ 
        & $\cO(dT^2)$ 
        & $\cO(dT^2)$
        \\ \hline
        \citet{chen2017zoo} 
        & $\cO\left({dn}/{\varepsilon^4}\right)$ 
        & $\cO(dnT)$ 
        & $\cO(dnT)$ 
        & None 
        & None 
        & None 
        \\ \hline
        \citet{huang2020accelerated} 
        & $\cO\left({dn^{1/2}}/{\varepsilon^2}\right)$ 
        & $\cO(nd)$ 
        & $\cO(n^{1/2}d)$ 
        & None 
        & None 
        & None 
        \\ \hline
        Our Method 
        & $\cO\left({d^{3/2}n^{1/2}}/{\varepsilon^2}\right)$ 
        & $\cO\left(nd^{1/2}\right)$ 
        & $\cO\left(n^{1/2}d^{1/2}\right)$ 
        & $\cO\left({dn^{1/2}}/{\varepsilon}\right)$ 
        & $\cO(n)$ 
        & $\cO(1)$ 
        \\ \hline
    \end{tabular}
\end{table}
\vspace{-9mm}

Zeroth-order Frank-Wolfe methods have gained significant attention due to their ability to optimize constrained problems without requiring explicit gradient information. 
In this section, we review key developments in zeroth-order Frank-Wolfe algorithms and highlight their limitations.

\citet{balasubramanian2018zeroth} proposed the zeroth-order stochastic conditional gradient method, 
which samples a function \( f_i(x) \) from the \( n \) functions constituting the objective \( f(x) \) in Eq.~\eqref{eq:prob}. 
Then one constructs an approximate gradient $\hat{\nabla} f_i(x, U,\mu)$ as Eq.~\eqref{eq:nab_h} where $U\in\RR^{d\times b}$ is a $d\times b$ random Gaussian matrix and $b$ is the sample size. 
Given the approximate gradient $\hat{\nabla} f_i(x, U,\mu)$, the standard Frank-Wolfe steps are conducted to update $x$.
To address the variance introduced by zeroth-order oracles, the algorithm requires \( b = \cO(dT) \), where \( T \) is the number of iterations, 
random directions per iteration. 
This high per-iteration query requirement renders the method impractical for high-dimensional problems. 
Additionally, due to the lack of variance reduction for stochastic sampling, the method suffers from low convergence rates, 
achieving query complexities of \( \cO(d / \varepsilon^4) \) and \( \cO(d / \varepsilon^3) \) for non-convex and convex functions, respectively, 
both of which are highly dependent on the target precision \( \varepsilon \).

To reduce the high query cost associated with \( \cO(dT) \) function evaluations, \citet{sahu2019towards} proposed sampling a single random direction to construct the approximate gradient. 
Although this approach significantly lowers the per-iteration query requirement to \( \cO(1) \), 
the method achieves a higher overall query complexity of \( \cO(d^{4/3} / \varepsilon^4) \) for non-convex functions, 
which is inferior to that of \citet{balasubramanian2018zeroth}.

\citet{huang2020accelerated} addressed the variance introduced by stochastic sampling by incorporating a variance reduction technique. 
This method achieves an improved query complexity of \( \cO(dn^{1/2} / \varepsilon^2) \) for non-convex functions. 
However, it does not effectively handle the variance caused by zeroth-order oracles, requiring an almost exact gradient estimation via \( 2nd \) queries to \( f(x) \) every \( \sqrt{n} \) iterations. 
Furthermore, the average per-iteration query cost remains \( \cO(n^{1/2} d) \), which limits its applicability in real-world high-dimensional problems.

In contrast, our framework introduces double variance reduction techniques to simultaneously address the variance induced by zeroth-order gradient estimators and stochastic sampling. 
For non-convex functions, our method requires at most \( \cO(nd^{1/2}) \) queries for certain iterations and an average of \( \cO(n^{1/2}d^{1/2}) \) queries per iteration. 
While our query complexity of \( \cO(d^{3/2}n^{1/2} / \varepsilon^2) \) is slightly higher than that of \citet{huang2020accelerated}, 
our method significantly reduces the query requirement per iteration, making it more practical for large-scale problems. 
For convex functions, our algorithm achieves a lower query complexity of \( \cO(dn^{1/2} / \varepsilon) \) with an average of \( \cO(1) \) queries per iteration. 
Table~\ref{tb:aa} provides a detailed comparison of query complexities across these methods.


\subsection{Contributions}

The main contributions of this paper can be summarized as follows:

\begin{itemize}
    \item \textbf{Development of a zeroth-order stochastic Frank-Wolfe framework with double variance reduction.} 
   We propose a novel zeroth-order stochastic Frank-Wolfe framework specifically designed for constrained finite-sum optimization problems. 
   By introducing double variance reduction techniques, the algorithm effectively reduces the variance induced by zeroth-order gradient estimators and stochastic sampling. 
   This improvement ensures enhanced convergence efficiency and query complexity compared to existing methods, such as those by~\citet{sahu2019towards} and~\citet{huang2020accelerated}.
   \item \textbf{Theoretical analysis of convergence for both convex and non-convex objectives.} 
   For convex objectives, our algorithm achieves a zeroth-order query complexity of \( O(d\sqrt{n}/\varepsilon) \), 
   and for non-convex objectives, the query complexity is \( O(d^{3/2}\sqrt{n}/\varepsilon^2) \). 
   These complexities improve over state-of-the-art methods, particularly in avoiding the heavy dependency on dimensionality \( d \) while maintaining efficient variance reduction (see Table \ref{tb:aa}).
   \item \textbf{Scalability and applicability to high-dimensional and large-scale ML tasks.} 
   Our algorithm is specifically designed to avoid the computation of full gradients or high-precision gradient approximations at any iteration. 
   Instead, it relies on low-cost function queries, making it well-suited for high-dimensional problems and large sample sizes in modern ML applications (see Section \ref{sec:experiment}).
\end{itemize}

\section{Preliminary}\label{sec:pre}

\subsection{Notation and Assumption}

We introduce several standard notations and assumptions that will be used in this paper.
First, we assume that the constrained set $\cX$ is bounded. 
This assumption is almost standard in analyzing the Frank-Wolfe approach \citep{jaggi13}.
\begin{assumption}
	The set $\cX$ is convex and bounded with $R^2$ (i.e., $\forall x, x' \in \cX, \Vert x-x'\Vert^2 \leq R^2$).\label{assump:bounded set}
\end{assumption}

Next, we will assume that the objective function $f(x)$ is $L$-smooth and each $f_i(x)$ is $\hL$-smooth.
\begin{assumption}\label{assump:smoothness}
	The function $f$ is $L$-smooth, that is, for all $x, y\in \RR^d$, it holds that,
	\begin{equation}
		f(y) \leq f(x) + \dotprod{\nabla f(x), y-x} + \frac{L}{2}\norm{y - x}^2.
	\end{equation}
\end{assumption}
\begin{assumption}\label{ass:hL}
	Each $f_i(x)$ is $\hL$-smooth, that is, for all $x, y\in\RR^d$, it holds that
	\begin{equation}
		\norm{\nabla f_i(x) - \nabla f_i(y)} \leq \hL \norm{x - y}.
	\end{equation}
\end{assumption}

Note that if each $f_i(x)$ is $\hL$-smooth, then we can directly derive that $f(x)$ is $\hL$-smooth by Eq.~\eqref{eq:prob}.
Thus, we can obtain the fact that $L\leq \hL$. 
In fact, $\hL$ could be $n$ times larger than $L$ ($L \leq \hL\leq n \cdot L$) in some extreme cases (see Example \ref{Ex:L and HatL} in Appendix \ref{appendix:example}).

\begin{assumption}	\label{assump:convex}
	The function $f$ is convex, i.e., for any $x$, $y \in \cX$, we have
	\begin{align}
		f(x) \geq f(y) + \dotprod{\nabla f(y), x-y}. \label{eq:convexity}
	\end{align}
\end{assumption}
Note that the convexity assumption of $f(x)$ is only used in the analysis of the Frank-Wolfe algorithm for convex cases in Section~\ref{sss:cc}.


\subsection{Zeroth-Order Gradient Estimate}

The zeroth-order technique requires approximating a gradient by only accessing function values. 
There are several ways to estimate the gradient. 
One popular way is approximating the gradient of a function $h$ at $x$ by two accesses to function value as follows, with $u$ being a $d$-dimensional Gaussian random vector:
\begin{equation}
	\hat{\nabla} h(x, u, \mu) = \frac{h(x + \mu u) - h(x -\mu u)}{2\mu}{u},
\end{equation}
where $\mu>0$.
Given above approximate gradient and letting  $U$ be a $d\times \cb$ Gaussian random matrix, that is, $U_{i,j} \sim \cN(0, 1)$, then we can define a batch version to approximate gradient $\nabla h(x)$ as follows:
\begin{equation}\label{eq:nab_h}
	\hat{\nabla} h(x, U, \mu) = \frac{1}{\cb} \sum_{j=1}^\cb \hat{\nabla} h(x, U_{:,j}, \mu),
\end{equation}
where $U_{:,j}=(U_{1,j},\dots,U_{d,j})^\top$ is the $j$th column of $U$.
The following lemma shows that $\hat{\nabla} h(x, U, \mu)$ is a reasonable estimate of $\nabla h(x)$.

\begin{lemma}\label{lem:nab_h}
Given a function $h(x)$ satisfying Assumption~\ref{assump:smoothness}, that is, $h(x)$ is $L$-smooth, then the approximate gradient $\hat{\nabla} h(x, U, \mu)$ defined in Eq.~\eqref{eq:nab_h} has the following property
	\begin{equation}\label{eq:naba_hp}
		\hat{\nabla} h(x, U, \mu) = \frac{1}{\cb} UU^\top \nabla h(x) + \frac{1}{\cb}\sum_{j=1}^{\cb}\tau_h(x, U_{:,j}, \mu)\cdot U_{:,j}, 
	\end{equation}
	with $\tau_h(x, U_{:, j}, \mu) = \frac{h(x + \mu U_{:, j}) - h(x-\mu U_{:, j}) -2\mu\dotprod{\nabla h(x), U_{:, j} }}{2\mu}$ and $|\tau_h(x, U_{:, j}, \mu)| \leq \frac{L\mu\norm{U_{:, j}}^2}{2}$.
\end{lemma}

Lemma~\ref{lem:nab_h} shows that the approximate gradient $\hat{\nabla} h(x, U, \mu)$ defined in Eq.~\eqref{eq:nab_h} can be decomposed into $\frac{1}{b}UU^\top \nabla h(x)$ and a bias term $\sum_{j=1}^b\tau_h(x, U_{:,j}, \mu)\cdot U_{:,j}$.
Note that we have  $\EE\left[\frac{1}{b}UU^\top \nabla h(x)\right] = \nabla h(x)$ and the bias term is of order $\cO(\mu)$.
Therefore, when $\mu$ goes to zero, $\hat{\nabla} h(x, U, \mu)$ can become an unbiased estimate of $\nabla h(x)$.

\subsection{Standard Frank-Wolfe Algorithm}

The standard Frank-Wolfe algorithm solves problem~\eqref{eq:prob} by the following iteration:
\begin{align}
	s_t &= \argmin_{s \in \cX}  \dotprod{s, \nabla f(x_t)}, \label{eq:lmo}\\
x_{t+1} &= (1 - \gamma_t) x_t + \gamma_t s_t, \label{eq:update}
\end{align}
where $\gamma_t \in (0, 1)$ is a step size.
In the Frank-Wolfe algorithm, a linear minimization (LMO) is conducted just as shown in Eq.~\eqref{eq:lmo}. 
The LMO can be computed efficiently in many applications, such as LASSO problem~\citep{tibshirani1996regression}. 
This is one of the reasons why the Frank-Wolfe algorithm has wide applications.

\section{ Zeroth-Order Stochastic Frank-Wolfe Framework}\label{sec:zosf_framework}

\subsection{Algorithm Development}

Our algorithm is built upon the standard Frank-Wolfe algorithm (Eq.~\eqref{eq:lmo}-\eqref{eq:update}), with specific modifications to address the challenges posed by the zeroth-order technique and the constrained finite-sum structure of problem~\eqref{eq:prob}. 
These challenges arise from two primary sources of variance: 

\begin{itemize}
    \item \textbf{Variance from zeroth-order gradient approximation:} Zeroth-order optimization relies on approximating gradients using function values, as shown in Eq.~\eqref{eq:nab_h}. 
    This approximation introduces inherent variance due to the estimation process. 
    Specifically, the variance scales as $\EE\left[\norm{\hat{\nabla} f(x, U, \mu) - \nabla f(x)}^2\right] = \cO\left(\norm{\nabla f(x)}^2\right)$~\citep{nesterov2017random}. 
    Unlike unconstrained problems, in constrained optimization, the gradient at the optimal point $x^*$ may not vanish ($\norm{\nabla f(x^*)} > 0$), leading to persistent variance that could not diminish to zero over iterations.
    \item \textbf{Variance from stochastic sampling in finite-sum problems:} In ML tasks, the sample size $n$ is often large, necessitating the use of stochastic optimization methods for computational efficiency. 
    However, these methods introduce additional variance due to random sampling, further impacting convergence rates \citep{johnson2013accelerating,bottou2010large}.
\end{itemize}

To address these dual sources of variance, we propose an enhanced zeroth-order stochastic Frank-Wolfe framework that employs double variance reduction techniques, targeting both the gradient approximation and the stochastic sampling variance by the following terms.

\begin{itemize}
    \item \textbf{Refined gradient estimation:} To mitigate the variance induced by zeroth-order gradient approximation, we introduce a refined gradient estimator $g_t$, inspired by variance reduction methods \citep{johnson2013accelerating,li2021page,hanzely2018sega}.
    The update rule is given by:
\begin{equation}\label{eq:g_def}
g_{t+1} = g_t + \frac{\cb}{d+\cb+1} \hat{\nabla} f(x_{t+1}, U_{t+1}, \mu_{t+1}) - \frac{U_{t+1} U_{t+1}^\top}{d+\cb+1} g_t,
\end{equation}
where $\hat{\nabla} f(x_{t+1}, U_{t+1}, \mu_{t+1})$ is the zeroth-order gradient approximation computed using Eq.~\eqref{eq:nab_h}. 
Here, $g_{t+1}$ captures the difference between the estimated gradient $\hat{\nabla} f(x_{t+1}, U_{t+1}, \mu_{t+1})$ and the projection of $g_t$ along the sampled direction $U_{t+1}$, thereby reducing variance from zeroth-order oracles~\citep{sahu2019towards,huang2020accelerated}.
In Section~\ref{sec:mail_proof}, we show that this approach ensures the variance of $g_t$ vanishes over iterations (See Lemmas \ref{lem:ggg} and \ref{lem:var}).

    \item \textbf{Variance reduction for finite-sum problems:} To further exploit the finite-sum structure in problem~\eqref{eq:prob}, we employ a refined batch version of the \textbf{P}rob\textbf{A}bilistic \textbf{G}radient \textbf{E}stimator (PAGE) method~\citep{li2021page} for gradient estimation, which reduces the computational cost of gradient updates. 
    The update rule is:
    \begin{small}
\begin{equation}\label{eq:g_page}
g_{t+1} = 
\begin{cases}
    g_t + \frac{\cb}{d+\cb+1} \hat{\nabla} f(x_{t+1}, U_{t+1}, \mu_{t+1}) - \frac{U_{t+1}U_{t+1}^\top}{d+\cb+1}g_t  & \text{with probability } p, \\
    g_t + \frac{1}{|\cS_t|} \sum_{i_t \in \cS_t} \left( \hat{\nabla} f_{i_t}(x_{t+1}, U_{t+1}, \mu_{t+1}) - \hat{\nabla} f_{i_t}(x_t, U_{t+1}, \mu_{t+1}) \right) & \text{with probability } 1-p.
\end{cases}
\end{equation}\end{small}
Here, $\cS_t$ is a randomly sampled subset of indices from $\{1, \dots, n\}$ with replacement, and we further assume that for any $t$ the sample size $|\cS_t|$ are the same, which is denoted as $|\cS|$. 
Compared with directly using $\hat{\nabla} f(x_{t+1}, U_{t+1}, \mu_t)$ to replace the full gradient $\nabla f(x_{t+1})$ in the standard PAGE method~\citep{li2021page}, we utilize Eq.\eqref{eq:g_def} due to the low zeroth-order estimate variance (see Lemma~\ref{lem:ggg}). 
\end{itemize}

With the gradient estimator $g_t$, the descent direction $s_t$ is determined by solving the linear minimization subproblem:
\begin{equation}\label{eq:LMO}
s_t = \argmin_{s \in \cX} \dotprod{s, g_t}.
\end{equation}
The next iterate is then updated as:
\begin{equation*}
x_{t+1} = (1 - \gamma_t) x_t + \gamma_t s_t,
\end{equation*}
where $\gamma_t \in (0, 1)$ is the step size.

In summary, our proposed algorithm integrates two layers of variance reduction to address the unique challenges of zeroth-order optimization in constrained finite-sum settings. 
By reducing variance from both zeroth-order gradient approximation and stochastic sampling, our method achieves improved convergence rates and computational efficiency. 
The complete algorithm is detailed in Algorithm~\ref{alg:zo_fw_vr}.

\begin{algorithm}[t!]
	\caption{Zeroth order Stochastic Frank-Wolfe with Double Variance Reduction}
	\label{alg:zo_fw_vr}
	\begin{small}
		\begin{algorithmic}[1]
			\STATE {\bf Input:}
			Input vector $x_0$, number iteration $T$, probability $p$;
			\STATE Compute $g_0 =  \hat{\nabla} f(x_0, U_0, \mu_0)$ with $U_0$ being a $d\times \cb$ Gaussian random matrix;
			\FOR{$t=0,\dots,T-1$}
			\STATE Compute $s_t = \argmin_{s\in \mathcal{S}} \dotprod{s, g_t}$; 
			\STATE Update $x_{t+1} = x_t + \gamma_t (s_t - x_t)$;
			\STATE Construct a $d\times \cb$ Gaussian random matrix $U_{t+1}$ and draw $z_t$ from the uniform distribution $U[0, 1]$;
			\IF{$z_t < p$}
			\STATE  Compute $\hat{\nabla} f(x_{t+1}, U_{t+1}, \mu_{t+1})$ as Eq.~\eqref{eq:nab_h};
			\STATE Update $g_{t+1} = g_t + \frac{\cb}{d+\cb+1} \hat{\nabla} f(x_{t+1}, U_{t+1}, \mu_{t+1}) - \frac{U_{t+1}U_{t+1}^\top}{d+\cb+1} g_t$;
			\ELSE  
			\STATE Generate a random set $\cS_t$ with size $|\cS_t|$;
			\STATE Update $g_{t+1} = g_t + \frac{1}{|\cS_t|} \sum_{i_t \in \cS_t} \left( \hat{\nabla} f_{i_t}(x_{t+1}, U_{t+1}, \mu_{t+1}) - \hat{\nabla} f_{i_t} (x_t, U_{t+1},\mu_{t+1})\right)$;
			\ENDIF
			\ENDFOR
			\STATE {\bf Return:}
			$x_{T}$.
		\end{algorithmic}
	\end{small}
\end{algorithm}


\vspace{-3mm}
\subsection{Convergence Analysis}

This section provides the theoretical analysis of Algorithm~\ref{alg:zo_fw_vr}. Specifically, we focus on two types of objective functions: convex and non-convex. 
For the convex case, we analyze the convergence properties of our algorithm and derive the complexity bounds in terms of zeroth-order queries and LMO calls. 
For the non-convex case, we establish convergence guarantees using the Frank-Wolfe gap as a criterion. 
We further derive the complexities of zeroth-order queries and LMO calls, highlighting the effectiveness of our double variance reduction approach even in non-convex scenarios.


\subsubsection{Convergence Analysis for Convex Objective}
\label{sss:cc}

\begin{theorem}\label{thm:main}
	Let Assumption~ \ref{assump:bounded set}-\ref{assump:convex} hold. Given the probability $p>0$ and sample size $|\cS|$, we define a Lyapunov function as follows:
	\begin{equation}\label{eq:lyapunov1}
		\Psi_t = f(x_t) - f(x^*) + \frac{2}{\sqrt{\frac{p\hL^2}{|\cS|} + 4pL^2}} \norm{g_t - \nabla f(x_t)}^2.
	\end{equation} 
	Let the step size $\gamma_t$ be properly chosen such that it satisfies, 
	\begin{equation}\label{eq:ss}
		\begin{aligned}
			&\text{if } T \leq \frac{8(d+\cb+1)}{p\cb}, &\gamma_t &= \frac{p\cb}{8(d+\cb+1)},\\
			&\text{if } T > \frac{8(d+\cb+1)}{p\cb}  \text{ and } t< t_0, &\gamma_t &= \frac{p\cb}{8(d+\cb+1)},\\
			&\text{if } T > \frac{8(d+\cb+1)}{p\cb} \text{ and } t \geq t_0, &\gamma_t &= \frac{2}{\frac{16(d+\cb+1)}{p\cb} + t - t_0},
		\end{aligned}
	\end{equation}
	where $t_0 = \lceil \frac{T}{2}\rceil$. 
	Choosing $\mu_t =  \sqrt{\frac{p\hL^2}{|\cS|} + 4pL^2} \cdot \frac{R\gamma_t}{(d+6)^{3/2}}$, then Lyapunov function defined as in Eq.~\eqref{eq:lyapunov1} with respect to Algorithm~\ref{alg:zo_fw_vr} satisfies the following property
	\begin{align}
		\EE\left[\Psi_{T+1}\right] \leq \frac{2^8(d+\cb+1)}{p\cb} \Psi_0 \exp\left(-\frac{Tp\cb}{16(d+\cb+1)}\right) + \frac{36\left( \frac{22(d+\cb+1)}{\cb} \cdot \sqrt{\frac{\hL^2}{p|\cS|} + \frac{4L^2}{p}} + \frac{L}{2} \right) \cdot R^2}{T}.\label{eq:zero_convergence}
	\end{align}
\end{theorem}

According to Theorem \ref{thm:main}, we derive the following corollary to show the LMO call and zeroth-order query complexities.

\begin{corollary}\label{cor:TQ}
	Under the conditions of Theorem~\ref{thm:main}, to find an $\varepsilon$-suboptimal solution,  Algorithm~\ref{alg:zo_fw_vr} requires to takes
	\begin{equation}\label{eq:T}
		T = \cO\left(\frac{d}{p\cb}\log\frac{1}{\varepsilon} + \frac{dR^2\sqrt{\frac{\hL^2}{p|\cS|} + \frac{L^2}{p}}}{\cb\varepsilon}\right) 
	\end{equation}
	LMO calls, that is, to solve the linear minimization in Eq.~\eqref{eq:LMO}. 
	Furthermore, our method takes 
	\begin{equation}\label{eq:Q}
		Q = \cO\left(\left(dn + \frac{d(1-p)}{p}\right)\log\frac{1}{\varepsilon} + \frac{dnR^2\sqrt{p} \sqrt{\frac{\hL^2}{|\cS|} + L^2}}{\varepsilon} + \frac{dR^2\sqrt{\frac{\hL^2|\cS|}{p} + \frac{L^2|\cS|^2}{p}}}{\varepsilon}\right)
	\end{equation}
	queries to function value $f_i(x)$'s.
\end{corollary}

\textbf{Proof:}
	To find an $\varepsilon$-suboptimal solution, the two terms in the right hand of Eq.~\eqref{eq:zero_convergence} being of order $\cO(\varepsilon)$ are sufficient.
	For the first term, we only need $T$ to satisfy
	\begin{align*}
		T = \frac{16(d+\cb+1)}{p\cb} \log \frac{2^9(d+\cb+1)}{p\cb \varepsilon} = \cO\left(\frac{d}{p\cb } \log\frac{1}{\varepsilon}\right).
	\end{align*}
	For the second term, we only need $T$ to satisfy
	\begin{align*}
		T = \frac{72\left( \frac{22(d+\cb+1)}{\cb} \cdot \sqrt{\frac{\hL^2}{p|\cS|} + \frac{4L^2}{p}} + \frac{L}{2} \right) \cdot R^2}{\varepsilon} 
		= \cO\left(\frac{dR^2\sqrt{\frac{\hL^2}{p|\cS|} + \frac{L^2}{p}}}{\cb\varepsilon}\right).
	\end{align*}
	Thus, Algorithm~\ref{alg:zo_fw_vr} takes 
	\begin{equation*}
		T = \cO\left(\frac{d}{p\cb } \log\frac{1}{\varepsilon} + \frac{dR^2\sqrt{\frac{\hL^2}{p|\cS|} + \frac{L^2}{p}}}{\cb\varepsilon}\right)
	\end{equation*}
	iterations to find an $\varepsilon$-suboptimal solution. 
	Since each iteration, Algorithm~\ref{alg:zo_fw_vr} takes an LMO call, the result in Eq.~\eqref{eq:T} has been proved.
	
	Furthermore, in these $T$ iterations, Algorithm~\ref{alg:zo_fw_vr} takes an update as Eq.~\eqref{eq:g_def} with a probability $p$ and takes an update as Eq.~\eqref{eq:g_page} with a probability $1 - p$.
	Thus, the total queries to function value $f_i(x)$'s is 
	\begin{align*}
		Q =& T\times p\times n\times \cb + T\times (1-p) \times \cb\times |\cS|\\
		  =& \cO\left(\left(dn + \frac{d(1-p)|\cS|}{p}\right)\log\frac{1}{\varepsilon} + \frac{dnR^2\sqrt{p} \sqrt{\frac{\hL^2}{|\cS|} + L^2}}{\varepsilon} + \frac{dR^2\sqrt{\frac{\hL^2 |\cS|}{p} + \frac{L^2|\cS|^2}{p}}}{\varepsilon}\right),
	\end{align*}  
	which has proved the result in Eq.~\eqref{eq:Q}.$\hfill\square$

Eq.~\eqref{eq:T} shows that a large probability $p$, $\cb$, and $|\cS|$ will effectively reduce the number of LMO calls. 
Unfortunately, this good property does not hold for the query complexity shown in Eq.~\eqref{eq:Q}. 
First, the query complexity $Q$ is independent of $\cb$.
Second, Eq.~\eqref{eq:Q} shows that $p$ and $|\cS|$ should be properly chosen to reach the minimum. 
It is easy to check that when $p = \frac{|\cS|}{n}$ and $|\cS| = 1$, Eq.~\eqref{eq:Q} reaches its minimum.

\begin{corollary}\label{cor:TQ1}
	Under the conditions of Theorem~\ref{thm:main}, to find an $\varepsilon$-suboptimal solution, Algorithm~\ref{alg:zo_fw_vr} with $p = \frac{|\cS|}{n}$ and $|\cS|=1$  requires
	\begin{equation*}
		T = \cO\left(\frac{d}{p\cb}\log\frac{1}{\varepsilon} + \frac{dR^2\sqrt{n\hL^2 + nL^2}}{\cb\varepsilon}\right) 
	\end{equation*}
	LMO calls. 
	Furthermore,  our method takes 
	\begin{equation}\label{eq:Q1}
		Q = \cO\left(dn\log\frac{1}{\varepsilon} + \frac{dR^2\sqrt{n} \sqrt{\hL^2 + L^2}}{\varepsilon} \right)
	\end{equation}
	queries to function value $f_i(x)$'s.
\end{corollary}

Theorem~\ref{thm:main} provides the convergence properties of our algorithm.
Corollary~\ref{cor:TQ} and \ref{cor:TQ1} give the detailed LMO and query complexities.
Specifically, when taking $p = \frac{1}{n}$ and $|\cS| = 1$ in Algorithm~\ref{alg:zo_fw_vr}, our algorithm can achieve a query complexity $\cO\left(dn\log\frac{1}{\varepsilon} + \frac{dR^2\sqrt{n} \sqrt{\hL^2 + L^2}}{\varepsilon} \right)$ which is the \emph{best} known query complexity of zeroth-order Frank-Wolfe algorithms for the finite-sum functions compared with state-of-the-art approaches (see Table \ref{tb:aa}). 

\subsubsection{Convergence Analysis for Non-Convex Objective}

Next, we will give the convergence property of our algorithm when the objective function $f(x)$ is non-convex.
Instead of using the Lyapunov's function defined in Eq.~\eqref{eq:lyapunov1} to describe the convergence property, we use the Frank-Wolfe gap function \citep{jaggi13} as a criterion for convergence:
\begin{equation*}
	\mathrm{gap}(y) = \max_{x\in\cX}\dotprod{\nabla f(y), y-x}.
\end{equation*}
Such a criterion is standard in the analysis of algorithms for the constrained problems with non-convex objective functions \citep{lacoste2016convergence}. It is easy to check that $\mathrm{gap}(y)\geq 0$ for
any $y \in \cX$. Moreover, a point $y\in \cX$ is stationary for problem~\eqref{eq:prob} if and only if $\mathrm{gap}(y)\geq 0$. 
\citet{lacoste2016convergence} notes that the Frank-Wolfe gap is a meaningful measure of non-stationary and also an
affine invariant generalization of the more standard convergence criterion $\norm{\nabla f(y)}$ that is used for unconstrained non-convex problems.

\begin{theorem}\label{thm:main1}
	Let Assumption~\ref{assump:bounded set}-\ref{ass:hL} hold. We also suppose Assumption~\ref{assump:bounded set} hold.
	Let $\{x_t\}_{t\geq 0}$ be generated by Algorithm~\ref{alg:zo_fw_vr} with $\gamma_t = \frac{1}{\sqrt{T}}$ and $\mu_t = \sqrt{\frac{p}{|\cS|(d+6)^3 T}} R$. 
	Let $x^*$ be the minimizer of the objective function with $f(x^*) > -\infty$.
	Then, Algorithm~\ref{alg:zo_fw_vr} has the following convergence property:
	\begin{equation}\label{eq:oo}
		\begin{aligned}
			&\frac{1}{T}\sum_{t=0}^{T-1}\mathrm{gap}(x_t) = \frac{1}{T}\sum_{t=0}^{T-1} \EE\left[\max_{x\in\cX}\dotprod{\nabla f(x_t), x_t-x}\right]\\ 
			\leq&
		\frac{f(x_0) - f(x^*)}{\sqrt{T}} +  \frac{8(d+\cb+1)R \norm{g_0 - \nabla f(x_0)}}{p\cb T} \\
		+& \sqrt{\frac{32(d+\cb+1)^2L^2}{p\cb^2} + \frac{40(d+\cb+1)^2\hL^2}{p|\cS|\cb^2}}\cdot \frac{R^2}{\sqrt{T}} + \frac{LR^2}{2\sqrt{T}}.
		\end{aligned}
	\end{equation}
\end{theorem}

According to Theorem \ref{thm:main1}, we derive the following corollary to show the LMO call and zeroth-order query complexities.

\begin{corollary}\label{cor:nonconv}
	Under the conditions of Theorem~\ref{thm:main1}, to find an $\varepsilon$-suboptimal solution,  Algorithm~\ref{alg:zo_fw_vr} with $p = \frac{1}{\sqrt{n}}$, $|\cS| = \sqrt{n}$, and $\cb = \sqrt{d}$ requires
	\begin{equation}\label{eq:T1}
		T = \cO\left(\left[\frac{f(x_0) - f(x^*)}{\varepsilon}\right]^2 + \frac{\sqrt{nd}\norm{g_0 - \nabla f(x_0)}R}{\varepsilon} + \frac{\sqrt{n} dL^2R^4}{\varepsilon^2} + \frac{d\hL^2R^4}{\varepsilon^2}\right) 
	\end{equation}
	LMO calls. Furthermore,  our method takes 
	\begin{equation}\label{eq:Q11}
		Q = \cO\left(\frac{\sqrt{nd}\big(f(x_0) - f(x^*)\big)^2}{\varepsilon^2} + \frac{n d^{3/2} L^2 R^4}{\varepsilon^2} + \frac{\sqrt{n}d^{3/2}\hL^2 R^4}{\varepsilon^2}\right)
	\end{equation}
	queries to function value $f_i(x)$'s. 
    Specifically, if $L\leq \hL n^{-1/4}$, the query complexity of our method reduces to 
    \begin{equation}\label{eq:Q12}
		Q = \cO\left(\frac{\sqrt{nd}\big(f(x_0) - f(x^*)\big)^2}{\varepsilon^2} +  \frac{\sqrt{n}d^{3/2}\hL^2 R^4}{\varepsilon^2}\right).
	\end{equation}
\end{corollary}

\textbf{Proof:} To find an $\varepsilon$-suboptimal solution, the four terms in the right hand of Eq.~\eqref{eq:oo} being of order $\cO(\varepsilon)$ are sufficient.
Thus, we only need $T$ to satisfy
\begin{align*}
	T =& \cO\left(\frac{(f(x_0) - f(x^*))^2}{\varepsilon^2} + \frac{\sqrt{nd}\norm{g_0 - \nabla f(x_0)}R}{\varepsilon} + \frac{\sqrt{n} dL^2R^4}{\varepsilon^2} + \frac{d\hL^2R^4}{\varepsilon^2}\right)\\
	=& \cO\left( \frac{(f(x_0) - f(x^*))^2}{\varepsilon^2} +  \frac{\sqrt{n} dL^2R^4}{\varepsilon^2} + \frac{d\hL^2R^4}{\varepsilon^2}\right),
\end{align*}
where the second equality is because  the term of order $\varepsilon^{-1}$ is a high order term.

Furthermore, in these $T$ iterations, Algorithm~\ref{alg:zo_fw_vr} takes an update as Eq.~\eqref{eq:g_def} with a probability $p$ and takes an update as Eq.~\eqref{eq:g_page} with a probability $1 - p$.
Thus, the total queries to function value $f_i(x)$'s is 
\begin{align*}
	Q =& T\times p\times n\times \cb + T\times (1-p) \times \cb\times |\cS|\\
	=&T\times \frac{1}{\sqrt{n}}\times n\times \sqrt{d} + T\times (1-\frac{1}{\sqrt{n}}) \times \sqrt{d}\times \sqrt{n}\\
	=& \cO\left(\frac{\sqrt{nd}\big(f(x_0) - f(x^*)\big)^2}{\varepsilon^2} + \frac{n d^{3/2} L^2 R^4}{\varepsilon^2} + \frac{\sqrt{n}d^{3/2}\hL^2 R^4}{\varepsilon^2}\right),
\end{align*}  
which has proved the result in Eq.~\eqref{eq:Q11}. If  $L\leq \hL n^{-1/4}$, then the third term of the right hand  of Eq.~\eqref{eq:Q11} dominates the complexity which implies Eq.~\eqref{eq:Q12}. $\hfill\square$

Corollary~\ref{cor:nonconv} shows that with $\cO(\sqrt{nd})$ queries to function value $f_i(x)$'s, our algorithm can achieve a query complexity $\cO(\frac{n d^{3/2} L^2 R^4}{\varepsilon^2} + \frac{\sqrt{n}d^{3/2}\hL^2 R^4}{\varepsilon^2})$ which is inferior to $\cO\left( \frac{d\sqrt{n}\hL^2R^4}{\varepsilon^2} \right)$ shown in \citep{huang2020accelerated}.
However, the work of \citet{huang2020accelerated} requires the computation of $d$ partial derivatives for each $n$ iterations, which is prohibited for high-dimension problems.
Using the random Gaussian direction similar to our method, \citet{chen2020frank} propose a zeroth-order Frank-Wolfe algorithm which achieves a query complexity $\cO\left(\frac{dn}{\varepsilon^4}\right)$ which is much inferior to our result.

\section{Main Proof}~\label{sec:mail_proof}
This section will give a detailed convergence analysis of our algorithm for both convex and non-convex problems. 
Accordingly, we will prove Theorems~\ref{thm:main} and~\ref{thm:main1}.

\vspace{-2mm}
\subsection{Proof of Theorem \ref{thm:main}}
\label{subsec:convex}

\begin{lemma}\label{lem:f_recursion}
	Let the objective function $f(x)$ satisfy Assumption~\ref{assump:bounded set}-\ref{assump:convex}.
	Given $\alpha>0$, then updates in Algorithm~\ref{alg:zo_fw_vr} satisfy 
	\begin{equation} \label{eq:ff}
		f(x_{t+1}) - f(x^*) \leq  (1-\gamma_t) (f(x_t)-f(x^*)) +  \alpha \Vert g_t - \nabla f(x_t)\Vert^2 +\frac{\gamma_t^2R^2}{2\alpha} + \frac{L\gamma_t^2R^2}{2}.
	\end{equation}
\end{lemma}

Next, we will bound the variance of our refined gradient estimator $g_t$ in the following lemma.
\begin{lemma}\label{lem:ggg}
Letting Assumption~\ref{assump:bounded set}-\ref{ass:hL} hold and the refined gradient estimator $g_t$ update as Eq.~\eqref{eq:g_page}, then sequence $\{g_t\}$ satisfies the following property
	\begin{equation}\label{eq:ggg}
		\begin{aligned}
			&\EE\left[\norm{g_{t+1} - \nabla f(x_{t+1})}^2\right] 
			\leq 
			\left(1 - \frac{p\cb}{4(d+\cb+1)}\right) \norm{g_t - \nabla f(x_t)}^2 \\
			&+\left(\frac{6p(d+\cb+1)L^2}{\cb} + \frac{2(d+\cb+1)\hL^2}{\cb|\cS_t|} + \frac{2(d+1)L^2}{\cb}\right) R^2\gamma_t^2
			+\frac{8(d+\cb+1)(d+6)^3}{p\cb}\hL^2\mu_{t+1}^2.
		\end{aligned}
	\end{equation}
\end{lemma}

Eq.~\eqref{eq:ggg} shows that the variance of $g_t$ will reduce at a rate $1 - \frac{p\cb}{4(d+\cb+1)}$ for each iteration but with some perturbations. 
If these perturbation terms vanish as the algorithm runs, then the variance of $g_t$ will also disappear.
This property is the key to the success of our algorithm.

Based on Lemma~\ref{lem:f_recursion} and Lemma~\ref{lem:ggg}, we provide the detailed proof of Theorem~\ref{thm:main}.

\textbf{Proof of Theorem \ref{thm:main}:} First, we represent the Lyapunov function in Eq.~\eqref{eq:lyapunov1} as follows:
	\begin{equation*}
		\Psi_t = f(x_t) - f(x^*) + M \cdot \norm{g_t - \nabla f(x_t)}^2,
	\end{equation*}
	with $M$ being defined as
	\begin{align*}
		M = \frac{8(d+\cb+1)\alpha}{p\cb}, \qquad\mbox{with}\qquad \alpha = \frac{\cb}{4(d+\cb+1)}\cdot \frac{1}{\sqrt{\frac{\hL^2}{p|\cS|} + \frac{4L^2}{p}} }.
	\end{align*}
	Thus, by the update of Algorithm~\ref{alg:zo_fw_vr}, we have
	\begin{align*}
		\EE\left[\Psi_{t+1}\right] 
		&=
		\EE\left[f(x_{t+1}) - f(x^*)\right] + M \cdot\EE\left[\norm{g_{t+1} - \nabla f(x_{t+1})}^2\right]\\
		&\stackrel{\eqref{eq:ff}\eqref{eq:ggg}}{\leq}
		(1 -\gamma_t) \big(f(x_t) - f(x^*)\big)+ \alpha\norm{g_t - \nabla f(x_t)}^2 + \frac{\gamma_t^2R^2}{2\alpha} + \frac{L\gamma_t^2R^2}{2}\\
		&+\left(1 - \frac{p\cb}{4(d+\cb+1)}\right)\cdot M\norm{g_t - \nabla f(x_t)}^2 \\
		&+ M \left(\frac{6p(d+\cb+1)L^2}{\cb} + \frac{2(d+\cb+1)\hL^2}{\cb|\cS_t|} + \frac{2(d+1)L^2}{\cb}\right) R^2\gamma_t^2\\
		&+M\frac{8(d+\cb+1)(d+6)^3}{p\cb}\hL^2\mu_t^2\\
		&=
		(1 -\gamma_t) \big(f(x_t) - f(x^*)\big)+ \left(1 - \frac{p\cb}{4(d+\cb+1)} + \frac{\alpha}{M}\right)\cdot M\norm{g_t - \nabla f(x_t)}^2 \\
		& +   M \left(\frac{6p(d+\cb+1)L^2}{\cb} + \frac{2(d+\cb+1)\hL^2}{\cb|\cS_t|} + \frac{2(d+1)L^2}{\cb}\right) R^2\gamma_t^2 + \frac{\gamma_t^2R^2}{2\alpha} + \frac{L\gamma_t^2R^2}{2}\\
		&+M\frac{8(d+\cb+1)(d+6)^3}{p\cb}\hL^2\mu_t^2\\
		&=
		(1 -\gamma_t) \big(f(x_t) - f(x^*)\big)+ \left(1 - \frac{p\cb}{8(d+\cb+1)}\right)\cdot M\norm{g_t - \nabla f(x_t)}^2\\
		&+ \frac{8(d+\cb+1)\alpha}{p\cb}\left(\frac{6p(d+\cb+1)L^2}{\cb} + \frac{2(d+\cb+1)\hL^2}{\cb|\cS_t|} + \frac{2(d+1)L^2}{\cb}\right) R^2\gamma_t^2 + \frac{\gamma_t^2R^2}{2\alpha} + \frac{L\gamma_t^2R^2}{2}\\
		&+M\frac{8(d+\cb+1)(d+6)^3}{p\cb}\hL^2\mu_t^2\\
		&\leq
		(1 -\gamma_t) \big(f(x_t) - f(x^*)\big)+ \left(1 - \frac{p\cb}{8(d+\cb+1)}\right)\cdot M\norm{g_t - \nabla f(x_t)}^2\\
		&+\left(\frac{16(d+\cb+1)^2\hL^2}{\cb^2 p|\cS_t|} + \frac{64(d+\cb+1)^2L^2}{\cb^2 p}\right)\alpha R^2\gamma_t^2 + \frac{\gamma_t^2R^2}{2\alpha} + \frac{LR^2\gamma_t^2}{2} \\
		&+ \alpha\cdot\frac{64(d+\cb+1)^2(d+6)^3}{p^2\cb^2} \hL^2\mu_t^2,
	\end{align*}
	where the last equality is because of $	M = \frac{8(d+\cb+1)\alpha}{p\cb}$. 
	Replacing  $\alpha = \frac{\cb}{4(d+\cb+1)}\cdot \frac{1}{\sqrt{\frac{\hL^2}{p|\cS|} + \frac{4L^2}{p}} }$ and $\mu_t =  \sqrt{\frac{p\hL^2}{|\cS|} + 4pL^2} \cdot \frac{R\gamma_t}{(d+6)^{3/2}}$ to above equation, we can obtain that
	\begin{align*}
		\EE\left[\Psi_{t+1}\right] 
		&\leq
		(1 -\gamma_t) \big(f(x_t) - f(x^*)\big)+ \left(1 - \frac{p\cb}{8(d+\cb+1)}\right)\cdot M\norm{g_t - \nabla f(x_t)}^2\\
		&+\frac{4(d+\cb+1)}{\cb} \cdot \sqrt{\frac{\hL^2}{p|\cS|} + \frac{4L^2}{p}} \cdot R^2\gamma_t^2 + \frac{2(d+\cb+1)}{\cb} \cdot \sqrt{\frac{\hL^2}{p|\cS|} + \frac{4L^2}{p}} \cdot R^2\gamma_t^2\\
		&+\frac{16(d+\cb+1)(d+6)^3}{p^2\cb} \cdot  \frac{\hL^2}{\sqrt{\frac{\hL^2}{p|\cS|} + \frac{4L^2}{p}} }\mu_t^2\\
		&=
		(1 -\gamma_t) \big(f(x_t) - f(x^*)\big)+ \left(1 - \frac{p\cb}{8(d+\cb+1)}\right)\cdot M\norm{g_t - \nabla f(x_t)}^2 \\
		&+ \frac{22(d+\cb+1)}{\cb} \cdot \sqrt{\frac{\hL^2}{p|\cS|} + \frac{4L^2}{p}} \cdot R^2\gamma_t^2 + \frac{LR^2\gamma_t^2}{2}\\
		&\leq
		(1 - \gamma_t)\cdot \Psi_t + \left( \frac{22(d+\cb+1)}{\cb} \cdot \sqrt{\frac{\hL^2}{p|\cS|} + \frac{4L^2}{p}} + \frac{L}{2} \right) \cdot R^2\gamma_t^2,
	\end{align*}
	where the last inequality is because of the step size setting that $\gamma_t \geq \frac{p\cb}{8(d+\cb+1)}$.
	
	Finally, by Lemma~\ref{lem:stich} with $a= \frac{p\cb}{8(d+\cb+1)}$ and $c = \left( \frac{22(d+\cb+1)}{\cb} \cdot \sqrt{\frac{\hL^2}{p|\cS|} + \frac{4L^2}{p}} + \frac{L}{2} \right) \cdot R^2$, we can obtain the final result. $\hfill\square$

\vspace{-3mm}
\subsection{Proof of Theorem \ref{thm:main1}}
\label{subsec:nonconv}

\begin{lemma}\label{lem:nonconv}
Letting Assumption~\ref{assump:bounded set}-\ref{assump:smoothness} hold, the sequence $\{x_t\}$ is generated by Algorithm~\ref{alg:zo_fw_vr}, then it holds that
	\begin{equation}\label{eq:ggap}
		\gamma_t\dotprod{\nabla f(x_t), x_t-x} 
		\leq f(x_t) - f(x^*) - \Big(f(x_{t+1}) - f(x^*)  \Big) + \gamma_t R\cdot \norm{\nabla f(x_t) - g_t}  + \frac{LR^2\gamma_t^2}{2}.
	\end{equation}
\end{lemma}

Next, we will give the upper bound of the variance of $g_t$ in an explicit form given the step size $\gamma_t$ and $\mu_t$. 

\begin{lemma}\label{lem:var}
Let Assumption~\ref{assump:bounded set}-\ref{ass:hL} hold and the refined gradient estimator $g_t$ update as Eq.~\eqref{eq:g_page}. 	
By choosing $\gamma_t = \frac{1}{\sqrt{T}}$ and $\mu_t = \frac{\sqrt{3}R p}{2(d+6)^{3/2}\sqrt{\cb T}}$, then sequence $\{g_t\}$ satisfies the following property
	\begin{equation}\label{eq:g_nab_1}
		\begin{aligned}
			\EE\left[\norm{g_{t+1} - \nabla f(x_{t+1})}^2\right]
			\leq& \left(1 - \frac{p\cb}{4(d+\cb+1)}\right)^{t+1} \norm{g_0 - \nabla f(x_0)}^2\\
			+& \left(\frac{32(d+\cb+1)^2L^2}{p\cb^2} + \frac{40(d+\cb+1)^2\hL^2}{p|\cS|\cb^2} \right) \cdot \frac{R^2}{T}.
		\end{aligned}
	\end{equation}
\end{lemma}

Based on the above two lemmas, we can prove Theorem~\ref{thm:main1}.

\textbf{Proof of Theorem~\ref{thm:main1}}

	By Lemma~\ref{lem:nonconv} and Lemma~\ref{lem:var}, we can obtain that
\begin{align*}
&\frac{1}{\sqrt{T}}\sum_{t=0}^{T-1}\EE\left[\max_{x\in\cX}\dotprod{\nabla f(x_t), x_t-x}\right]\\
&\stackrel{\eqref{eq:ggap}}{\leq}
\sum_{t=0}^{T-1}\left[\EE\left[f(x_t) - f(x^*)\right] - \EE\left[f(x_{t+1}) - f(x^*)\right] + \frac{R}{\sqrt{T}} \cdot \EE\left[\norm{\nabla f(x_t) - g_t}\right]  + \frac{LR^2}{2T}\right]\\
&\leq
f(x_0) - f(x^*) - \EE\left[f(x_T) - f(x^*)\right] + \frac{R}{\sqrt{T}}\cdot \sum_{t=0}^{T-1}\sqrt{\EE\left[\norm{\nabla f(x_t) - g_t}^2\right]}  + \frac{LR^2}{2}\\
&\stackrel{\eqref{eq:g_nab_1}}{\leq}
f(x_0) - f(x^*) + \frac{R}{\sqrt{T}} \sum_{t=0}^{T-1} \left( 1 - \frac{p\cb}{8(d+\cb+1)} \right)^t \cdot  \norm{g_0 -\nabla f(x_0)} \\
&+ \frac{R}{\sqrt{T}} \cdot \sqrt{\frac{32(d+\cb+1)^2L^2}{p\cb^2} + \frac{40(d+\cb+1)^2\hL^2}{p|\cS|\cb^2}}\cdot \frac{R}{\sqrt{T}} + \frac{L^2R^2}{2}\\
&\leq
f(x_0) - f(x^*) + \frac{8(d+\cb+1)R \norm{g_0 - \nabla f(x_0)}}{p\cb\sqrt{T}} \\
&+ \frac{LR^2}{2} + \sqrt{\frac{32(d+\cb+1)^2L^2}{p\cb^2} + \frac{40(d+\cb+1)^2\hL^2}{p|\cS|\cb^2}}\cdot R^2.
\end{align*}	
Dividing both sides of above equation by $\sqrt{T}$, then we have
\begin{align*}
&\frac{1}{T}\sum_{t=0}^{T-1} \EE\left[\max_{x\in\cX}\dotprod{\nabla f(x_t), x-x_t}\right]
\leq
\frac{f(x_0) - f(x^*)}{\sqrt{T}} +  \frac{8(d+\cb+1)R \norm{g_0 - \nabla f(x_0)}}{p\cb T} \\
&+ \sqrt{\frac{32(d+\cb+1)^2L^2}{p\cb^2} + \frac{40(d+\cb+1)^2\hL^2}{p|\cS|\cb^2}}\cdot \frac{R^2}{\sqrt{T}} + \frac{LR^2}{2\sqrt{T}},
\end{align*}
which concludes the proof.$\hfill\square$

\section{Experiment}\label{sec:experiment}

In this section, we present a comprehensive evaluation of the proposed zeroth-order stochastic Frank-Wolfe algorithm (denoted as \textit{ZSFW-DVR}) through three sets of experiments on representative ML tasks: black-box sparse logistic regression, black-box robust classification, and black-box adversarial attacks.
Across all experiments, the proposed method is bench-marked against state-of-the-art baseline algorithms to evaluate its performance. 
\vspace{-2mm}
\subsection{Convex Case: Black-box Sparse Logistic Regression}
This subsection aims to demonstrate the effectiveness of the proposed algorithm based on the ML task, sparse logistic regression in the black-box setting.
Following the setting from \citet{liu2009large}, consider a dataset $\mathcal{D} = \{(z_i, y_i)\}_{i=1}^n$, where $z_i \in \mathbb{R}^d$ denotes the feature vector and $y_i \in \{-1, +1\}$ represents the corresponding binary label. 
The sparse logistic regression can be formulated as a finite-sum optimization as given in Eq.~\eqref{eq:logistic} as follows:
\begin{equation}\label{eq:logistic}
    \begin{split}
        &\min_{x} \, \frac{1}{n} \sum_{i=1}^{n} \log(1+\exp(-y_i x^\top z_i)),\\
        &\mbox{s.t.} \quad \Vert x\Vert_1 \leq r,
    \end{split}
\end{equation}
where $f_i(x) = \log(1+\exp(-y_i x^\top z_i))$ denotes the logistic loss for the $i$-th sample, and the constraint restricts $x$ to lie within the $\ell_1$-ball of radius $r$. 
In the black-box setting, direct access to the gradient of $f_i(x)$ is unavailable. 
Instead, the optimization process must rely on approximate gradient estimation techniques, such as $\hat{\nabla} f(x, U, \mu)$ defined in Eq.~\eqref{eq:nab_h}, which approximates the gradient using only function value queries. 


We conduct experiments on four publicly available datasets—a9a, RCV1, Real-Sim, and w8a—which exhibit diverse characteristics in terms of feature dimensionality (ranging from low to high) and class distribution (from balanced to highly imbalanced). 
We compare the performance of our algorithm ZSFW-DVR against three baseline methods:
(1) the deterministic zeroth-order Frank-Wolfe algorithm (ZOFWGD) (see Algorithm 1 of \citet{sahu2019towards}); 
(2) the stochastic zeroth-order Frank-Wolfe algorithm (ZOFWSGD) (see Algorithm 2 of \citet{sahu2019towards}); 
and (3) the accelerated deterministic zeroth-order Frank-Wolfe algorithm (AccSZOFW) proposed by \citet{huang2020accelerated}. 
To evaluate performance, we measure the objective gap, defined as $f(x) - f(x^*)$, as a function of query complexity. 

\begin{figure}[htp]
\centering
\subfloat[RCV1]{%
  \includegraphics[width=0.299\textwidth]{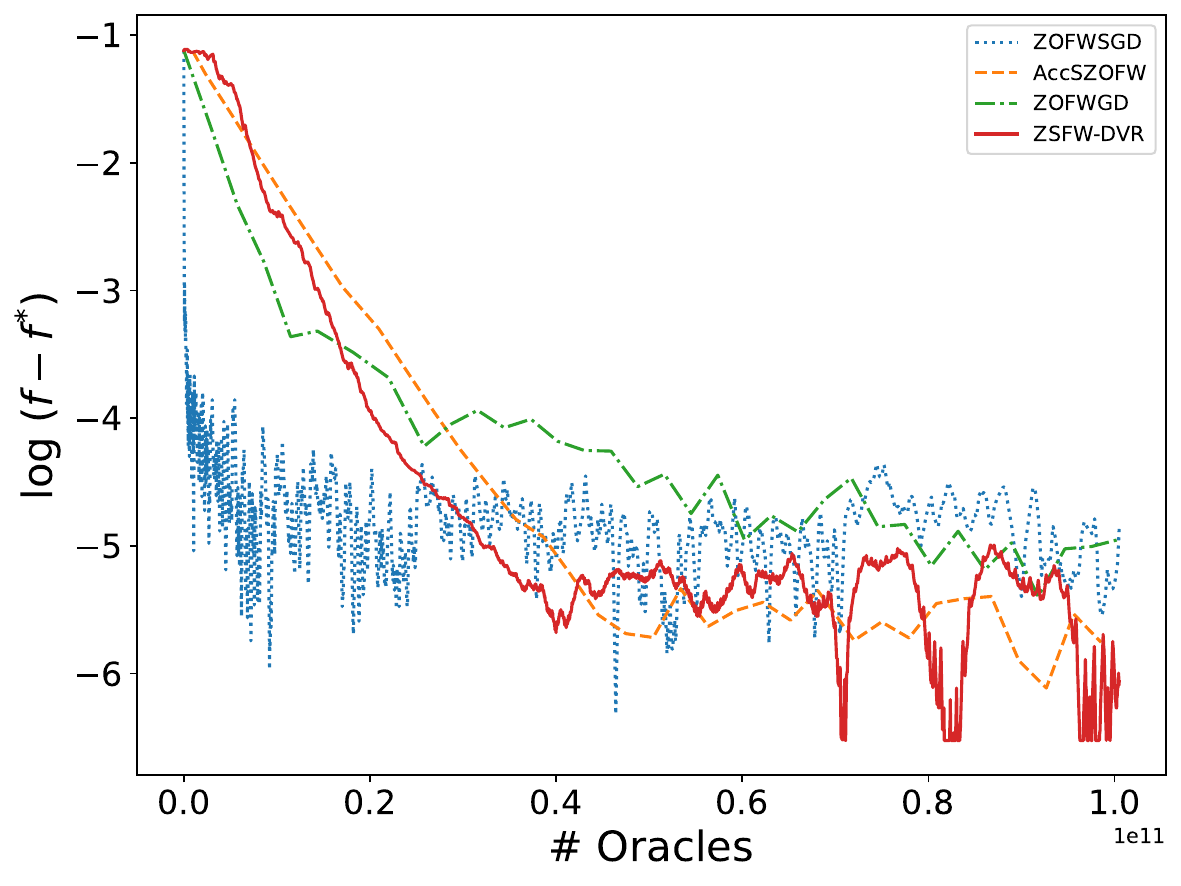}%
}
\subfloat[Real-Sim]{%
  \includegraphics[width=0.3\textwidth]{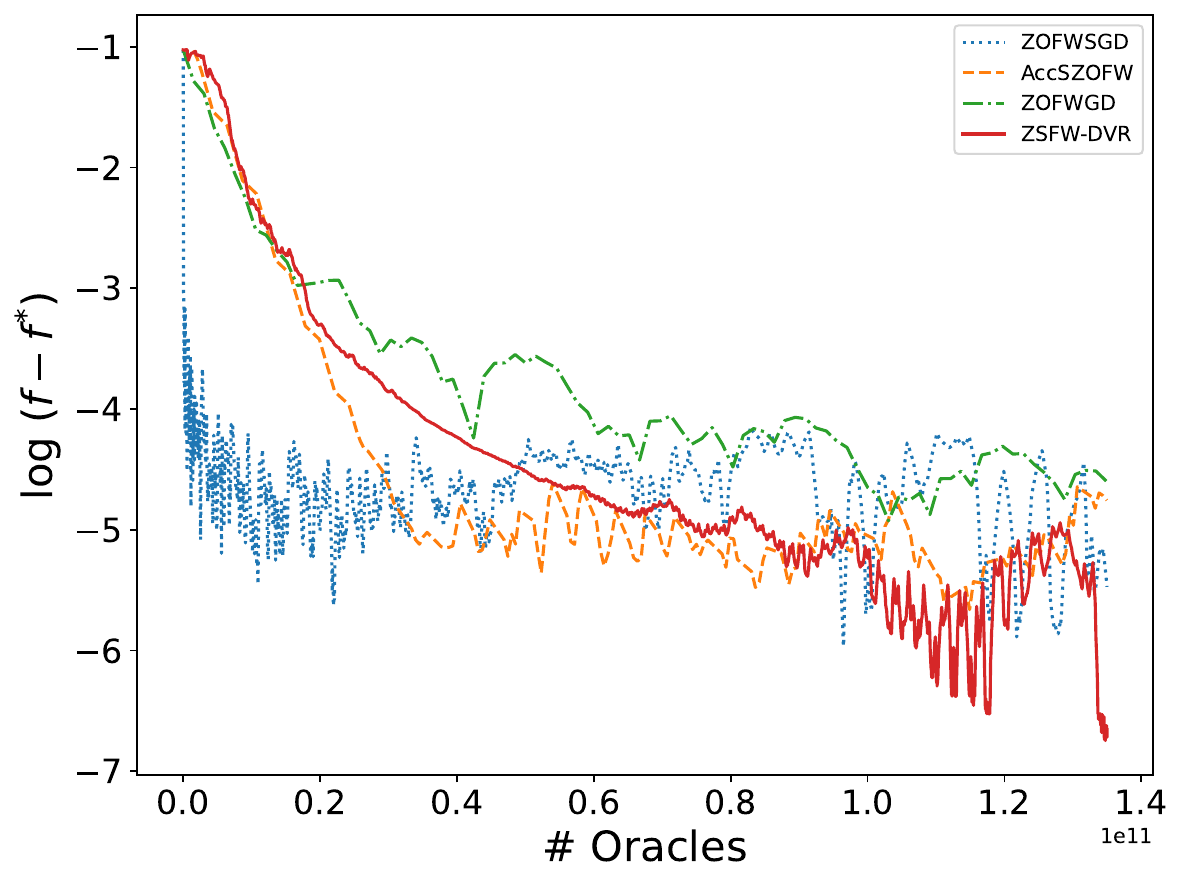}%
}%

\subfloat[a9a]{%
  \includegraphics[width=0.3\textwidth]{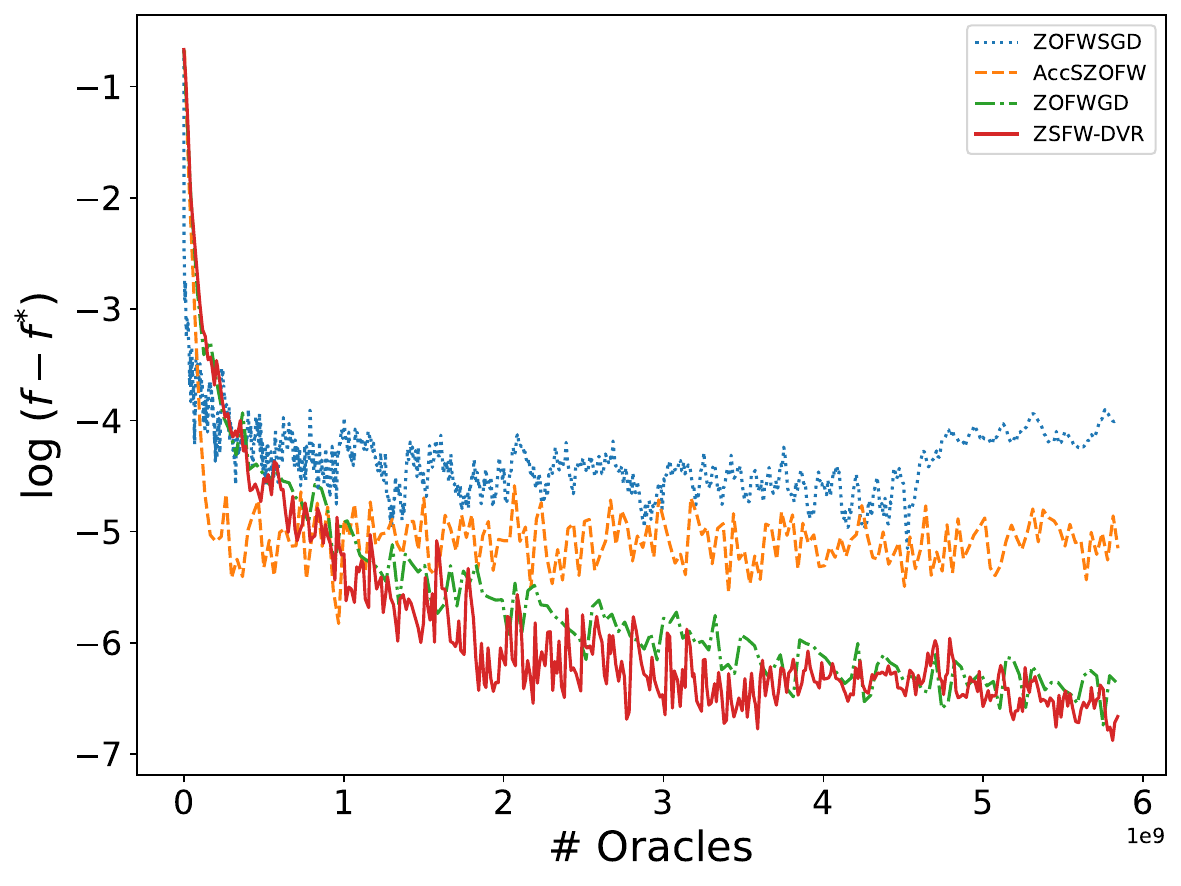}%
}%
\subfloat[w8a]{%
  \includegraphics[width=0.3\textwidth]{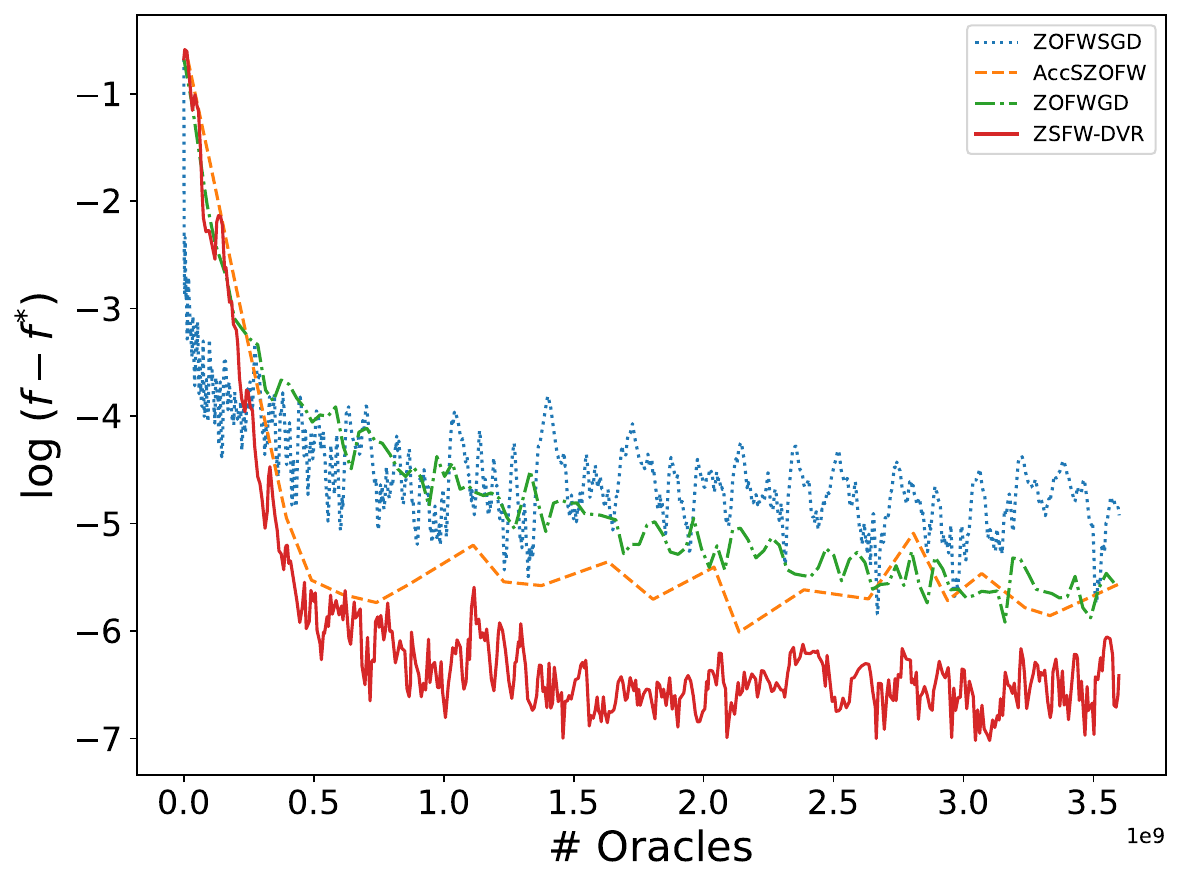}%
}


\caption{Objective gap comparison with different algorithms for black-box sparse logistic regression. The $y$ axis represents the logarithm (base 10) of the objective gap, and the $x$ is the number of queries during the optimization process.}
\label{fig:convex_result}
\end{figure}

As illustrated in Figure~\ref{fig:convex_result}, our proposed algorithm consistently achieves the smallest objective gap for a given number of queries across all four datasets. 
Although ZOFWSGD demonstrates faster initial convergence on three datasets, its stochastic nature, combined with the variance introduced by black-box gradient estimation, hinders its ability to reach an optimal solution, resulting in a significantly higher final objective gap. 
On the other hand, AccSZOFW utilizes coordinate-wise black-box gradient estimators, which necessitate a substantially larger number of queries to approximate the optimal solution. 
Despite employing acceleration techniques, AccSZOFW exhibits slower convergence on the RCV1 and w8a datasets, and its final objective gap remains consistently higher than that of our algorithm across all datasets. 
Similarly, ZOFWGD consistently underperforms compared to ZSFW-DVR in terms of both convergence rate and final objective gap, highlighting its limitations in handling high-dimensional and constrained optimization tasks.

\vspace{-5mm}
\subsection{Non-Convex Case: Black-box Robust Classification}


We consider a robust classification problem against a training dataset that contains outliers (e.g., mislabelled samples and inconsistent annotation standards).
We use the same setting from \citet{huang2020accelerated} to consider the robust classification with a correntropy-induced loss function. 
The robust classification task on a noisy dataset $\mathcal{D}=\{(z_i, y_i)\}_{i=1}^n$ can be formulated as a finite-sum optimization problem:
\begin{equation*}
    \begin{split}
        &\min_{x} \, \frac{1}{n}\sum_{i=1}^n 50 * (1-\exp(-\frac{(y_i - x^\top z_i)^2}{100})),\\
        &\mbox{s.t.} \quad \Vert x\Vert_1 \leq r,
    \end{split}
\end{equation*}
where $z_i \in \mathbb{R}^d$ is the feature vector, $y_i$ denotes the corresponding label, and $f_i(x)= 50 * (1-\exp(-\frac{(y_i - x^\top z_i)^2}{100}))$ represents the robust correntropy-induced loss for sample $i$.


\begin{figure}[htp]
\centering

\subfloat[RCV1]{%
  \includegraphics[width=0.3\textwidth]{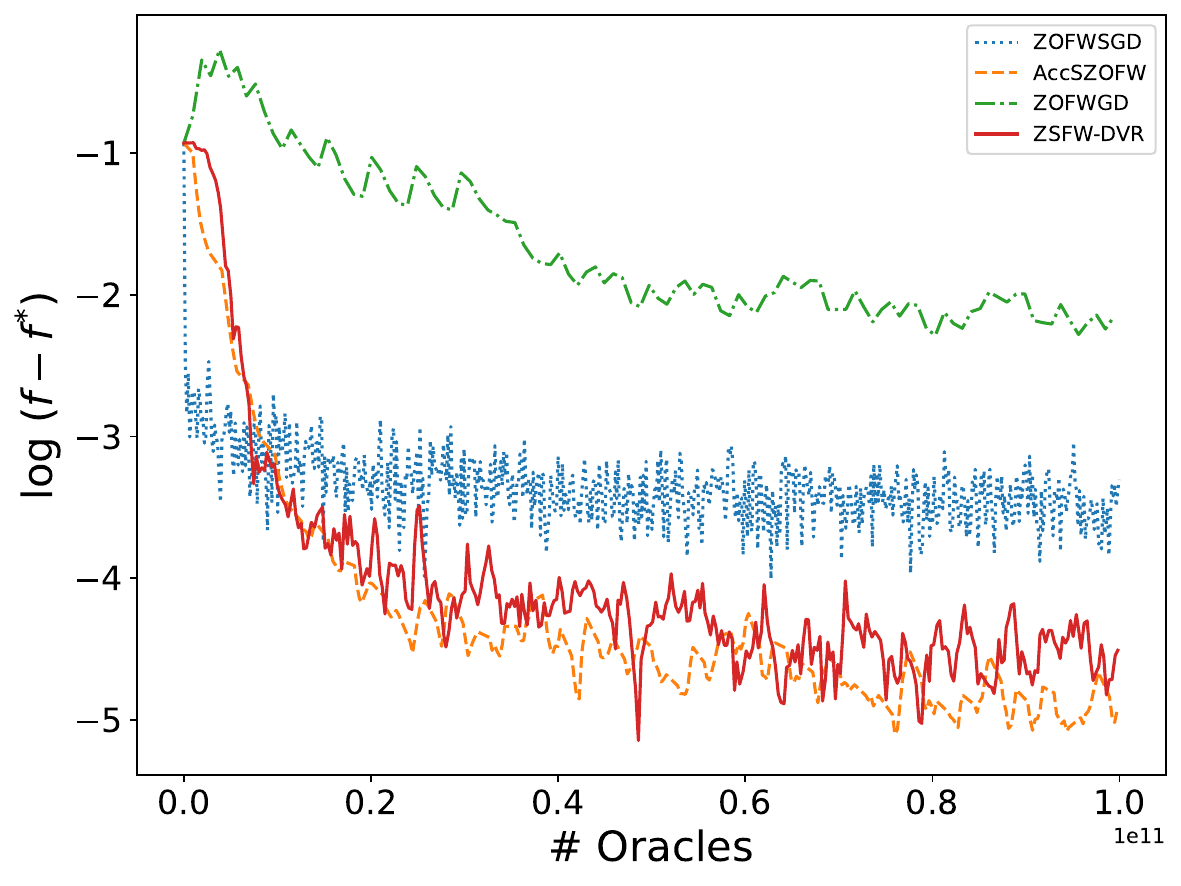}%
}
\subfloat[Real-Sim]{%
  \includegraphics[width=0.3\textwidth]{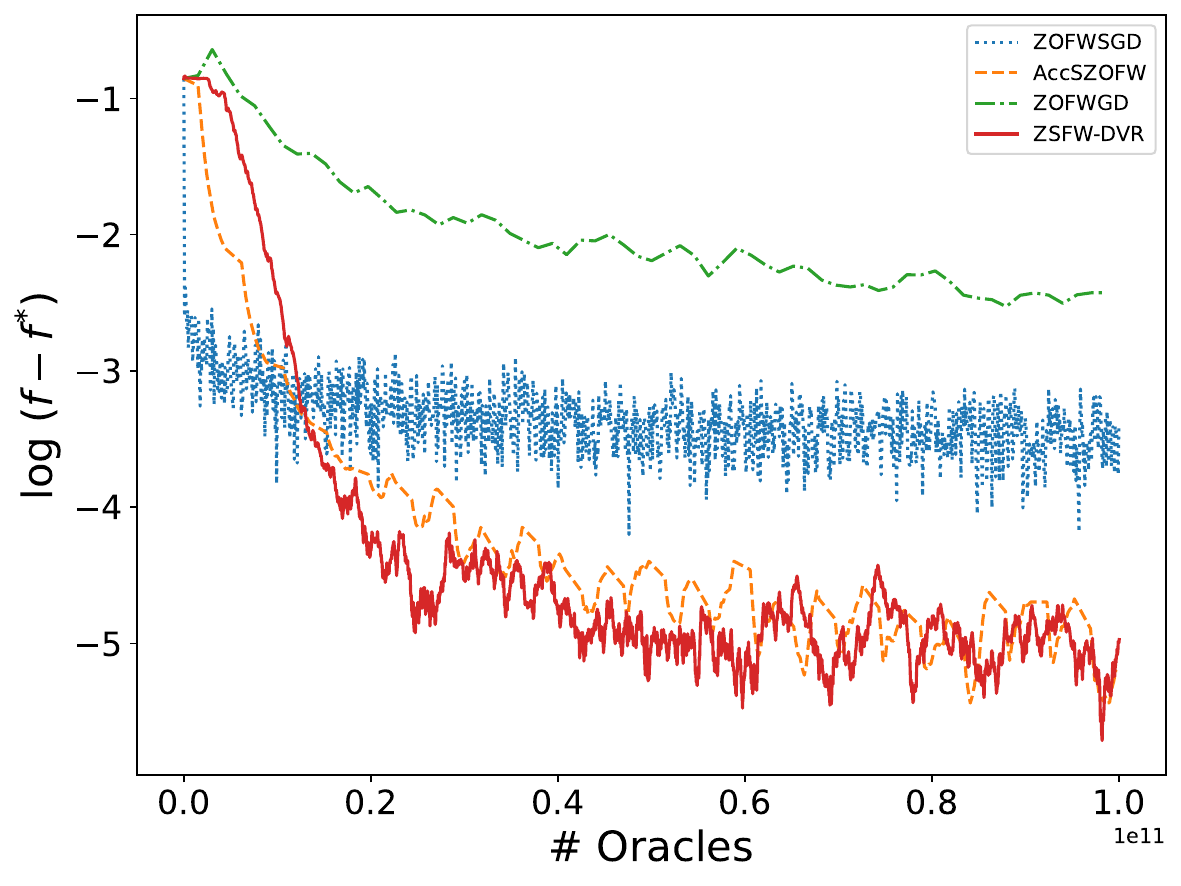}%
}%

\subfloat[a9a]{%
  \includegraphics[width=0.3\textwidth]{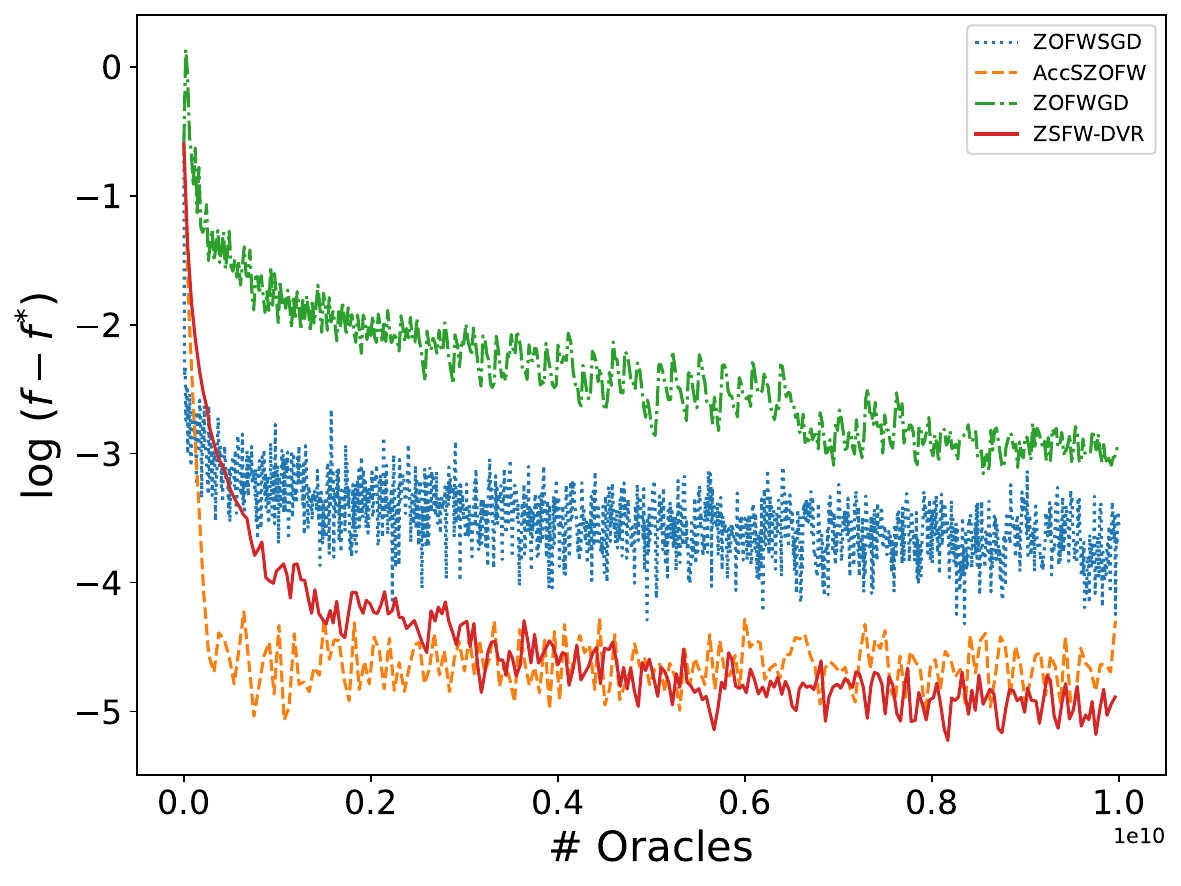}%
}%
\subfloat[w8a]{%
  \includegraphics[width=0.3\textwidth]{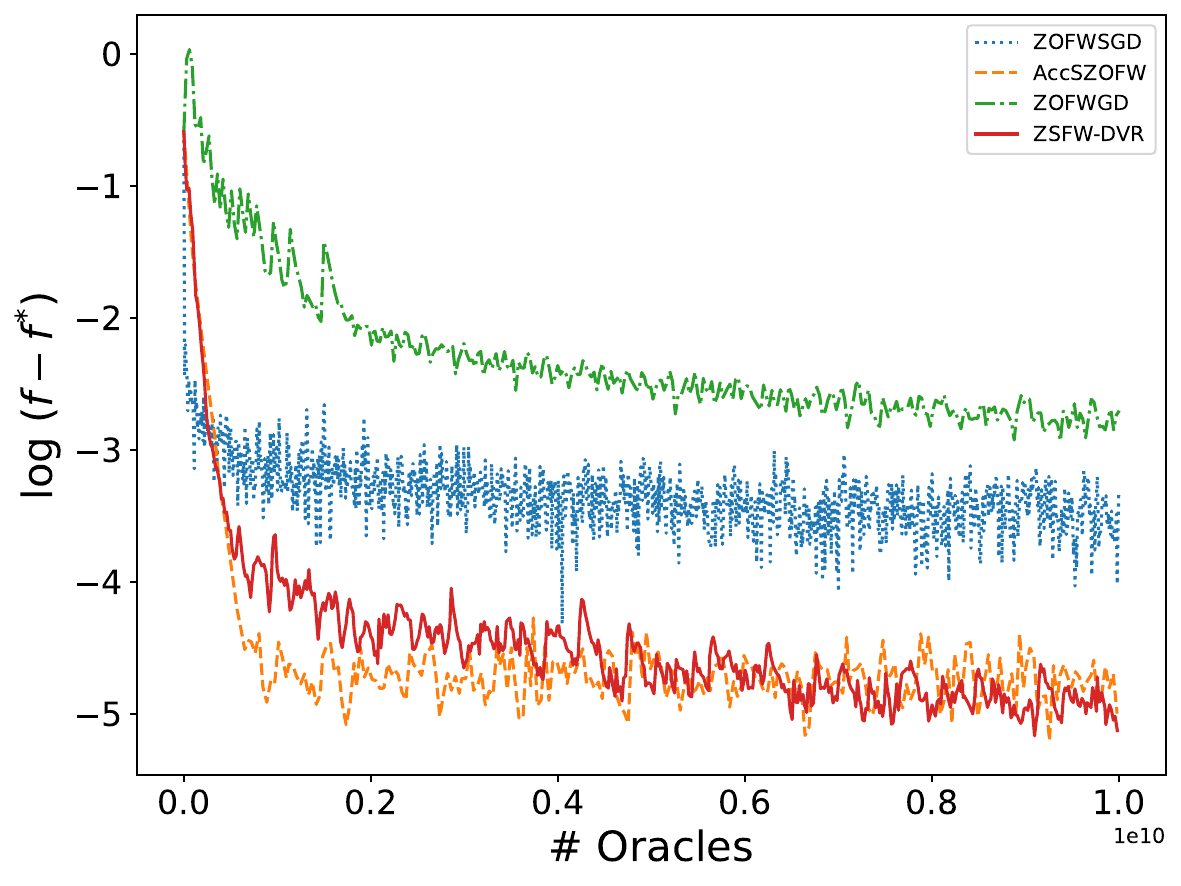}%
}

\caption{Objective gap comparison with different algorithms for black-box robust regression. The setting of the $x$ and $y$ axis are the same with Figure \ref{fig:convex_result}.}
\label{fig:nonconvex_result}
\end{figure}
\vspace{-1mm}

In this experiment, we evaluate the performance of our algorithm on black-box robust classification tasks using the same benchmark datasets as in the black-box sparse logistic regression experiment. We compare our method against (1) ZOFWGD; (2) the non-convex  stochastic zeroth-order Frank-Wolfe algorithm (see Algorithm 3 in~\citet{sahu2019towards}); (3) AccSZOFW.  
The detailed parameter settings are provided in Appendix~\ref{appendix:setting}.

The experiment results are presented in Figure \ref{fig:nonconvex_result}. 
First, ZSFW-DVR achieves the lowest objective gap on Real-Sim, a9a, and w8a, demonstrating its effectiveness in handling black-box robust classification tasks. 
On the RCV1 dataset, ZSFW-DVR performs slightly worse than the AccSZOFW method. This difference might stem from the higher dimensionality of the RCV1 dataset, as the query complexity for AccSZOFW scales as $\cO\left({dn^{1/2}}/{\varepsilon^2}\right)$, while our algorithm scales as $\cO\left({d^{3/2}n^{1/2}}/{\varepsilon^2}\right)$.  Notably, the ZOFWSGD method shows a fast initial decline in the objective gap but fails to make further progress after a certain number of iterations. This stagnation suggests that the noise introduced by zeroth-order gradient estimation can significantly hinder its ability to approach the optimal solution in later stages of optimization. In contrast, the ZOFWGD algorithm exhibits the slowest convergence rate and the highest objective gap across all datasets. Overall, ZSFW-DVR demonstrates superior performance in terms of both convergence speed and final objective value across most datasets, underscoring its advantage in tackling black-box robust classification problems.

\vspace{-2mm}

\subsection{Non-Convex Case: Black-box Constrained  Adversarial Attacks}

In the third experiment, we consider a universal adversarial attack whose goal is to construct a single perturbation that consistently misleads the target model into producing incorrect predictions when added to a diverse set of input samples~\citep{liu2018zeroth}.
More specifically, we focus on the constrained adversarial perturbation setting.
Let $\mathcal{D} = \{ (z_i, y_i)\}_{i=1}^{n}$ represent the image dataset, where $z_i$ denotes the $i$th original image with ground-truth label $y_i$. 
We assume $z_i \in [-0.5,0.5]^{d}$, which aligns with the standard preprocessing step of normalizing image data before feeding it into classification models. 
For each input image $z_i$, the target model outputs a prediction score vector $F(z_i) = [F_1(z_i), F_2(z_i),...,F_K(z_i) ]^\top$, where $K$ is the number of image classes. 
Each $F_j(z_i)$ indicates the model's confidence (e.g., log-probability or raw prediction score) that the image $z_i$ belongs to class $j$. 
The prediction label is given by $\hat{y_i} = \arg \max_j F_j(z_i)$.  We introduce the loss function: $f_i(x) = \log \bigg(   F_{y_i}(\frac{1}{2} \cdot \tanh ( \tanh^{-1} 2 z_i + x ))  \bigg) -   \log \bigg( \max_{j \neq y_i}    F_{j}(\frac{1}{2} \cdot \tanh ( \tanh^{-1} 2 z_i + x )) \bigg) $. Minimizing $f_i(x) $ aims to reduce the confidence margin, increasing the likelihood of misclassification under the perturbation $x$. 
Finally, the constrained universal adversarial perturbation problem can be formed as:
\begin{equation*}
    \begin{split}
        &\min_{x} \, \frac{1}{n} \sum_{i=1}^{n} f_i(x),\\
        &\mbox{s.t.} \quad \Vert x\Vert_2 \leq r,
    \end{split}
\end{equation*}
where $f_i(x)$ represents the adversarial loss for the $i$th image, and $r$ denotes the distortion budget for perturbation measured by the $\ell_2$-norm. 

We conduct experiments on two image datasets: MNIST \citep{lecun2010mnist} and CIFAR-10 \citep{krizhevsky2009cifar}. 
For the MNIST dataset, we employ a pre-trained deep neural network as the target black-box model, which achieves 99.4\% accuracy. 
We select 1131 correctly classified images of the digit ``1" as our target samples. 
For CIFAR-10, we pre-train a ResNet-18 classification model with a accuracy of 91.3\%. 
We select 45,628 correctly classified images from all 10 classes as our target samples, introducing a more diverse and challenging adversarial scenario. (The statistics about datasets and detailed parameter settings are provided in Appendix~\ref{appendix:setting}.)
In this scenario, we focus on the attack success rate (ASR) as our primary evaluation metric. The ASR, which measures the proportion of target samples successfully misclassified by the target model after applying the adversarial perturbation, provides a more practical measure of the effectiveness of an adversarial attack.

\begin{figure}[htp]
\centering
\subfloat[MNIST (Class 1)]{%
  \includegraphics[width=0.35\textwidth]{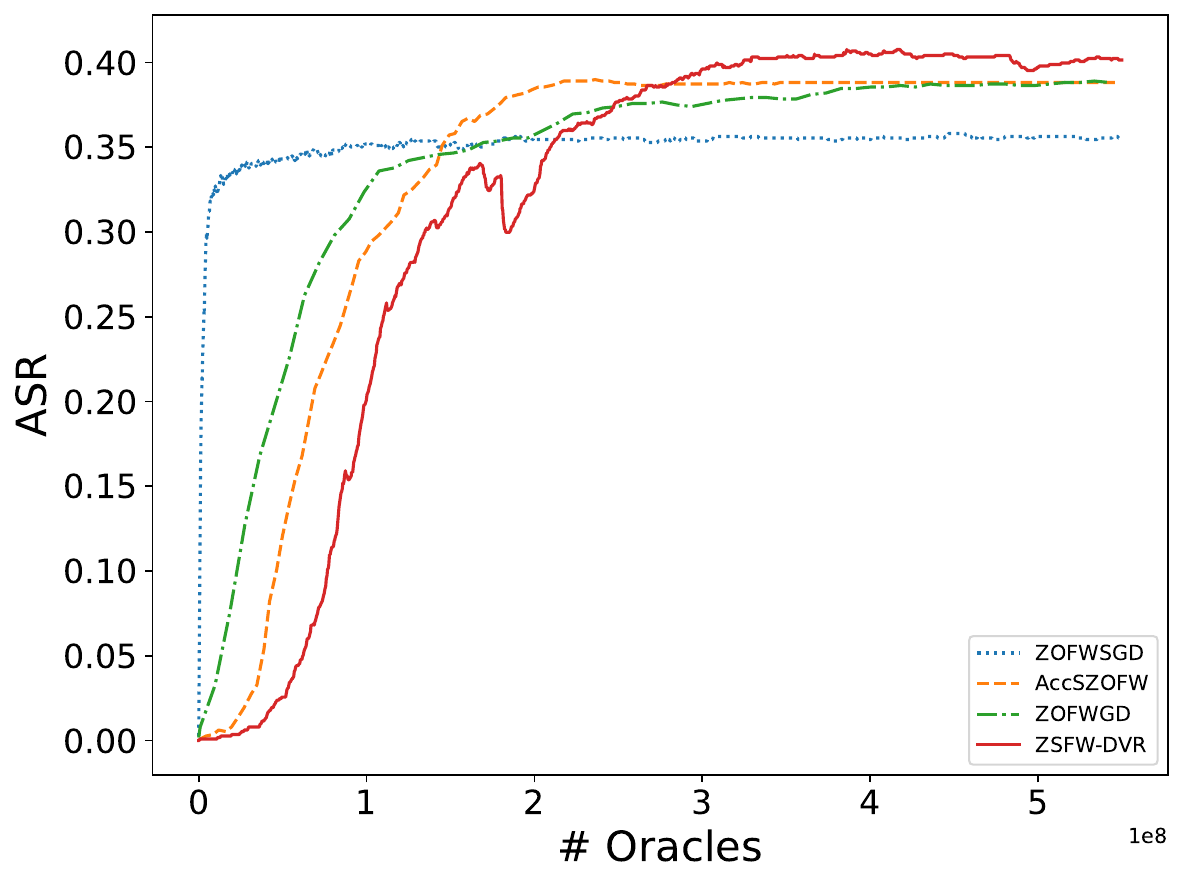}%
}%
\subfloat[CIFAR-10]{%
  \includegraphics[width=0.35\textwidth]{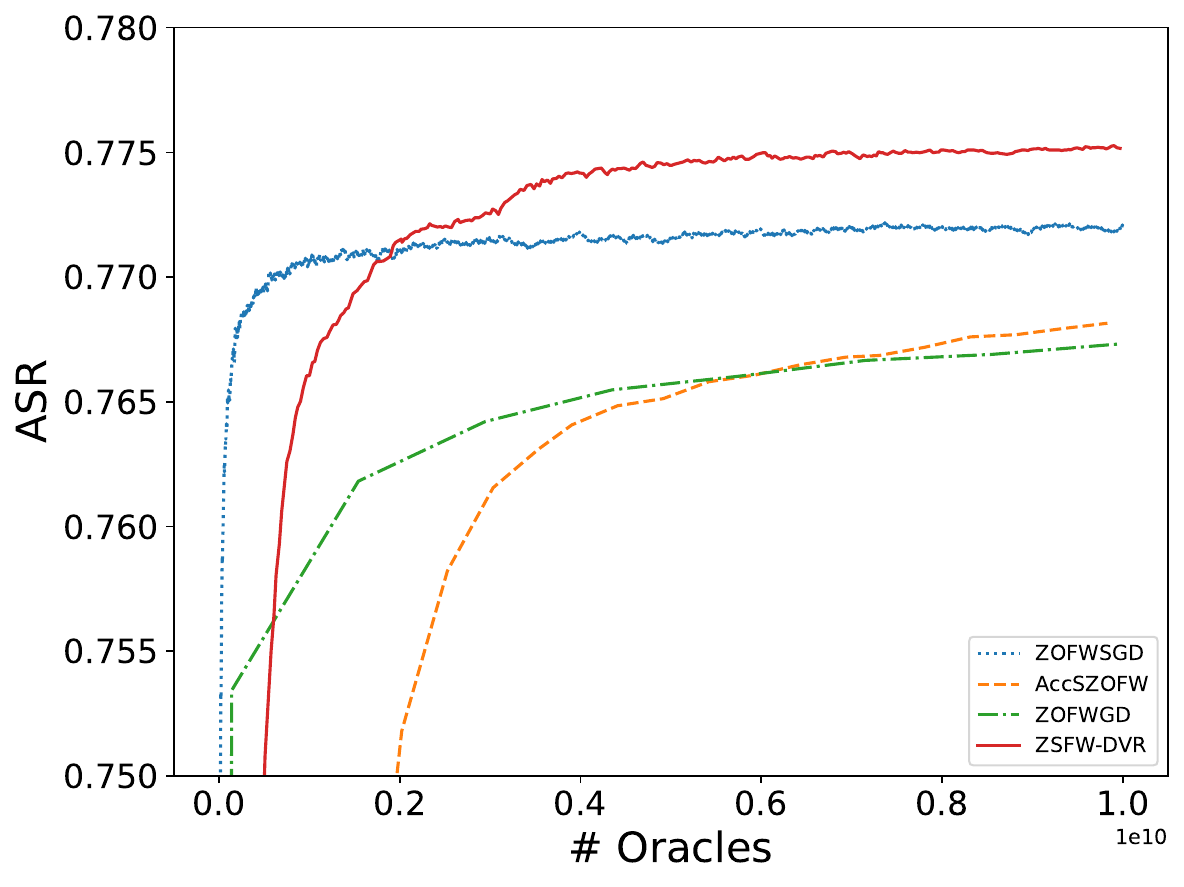}%
}%

\caption{Attack success rates against the number of queries on MNIST and CIFAR-10 datasets. }
\label{fig:img_succ_rate_trace}
\end{figure}

The attack success rates against the number of queries are depicted in Figure \ref{fig:img_succ_rate_trace}. Similarly, ZOFWSGD has rapid initial progress but stagnates prematurely in both datasets, highlighting its limitations in approaching optimal solutions. On MNIST, our method ZSFW-DVR achieves the highest attack success rate, with AccSZOFW closely following as the second-best performer for a given query budget. On CIFAR-10, which presents more significant challenges due to its larger number of target images and higher-dimensional input space, ZSFW-DVR also achieves the highest attack success rate with a given number of queries. 
These results demonstrate the effectiveness of ZSFW-DVR, particularly in handling higher-dimensional and more complex datasets. Additionally, we provide visual comparisons of original images and their adversarial examples generated by each algorithm in Appendix~\ref{appendix:setting} (see Figure \ref{fig:img_visual}).


\vspace{-5mm}

\section{Conclusion}\label{sec:conclusion}

This paper presents an enhanced zeroth-order stochastic Frank-Wolfe framework designed for constrained finite-sum optimization, a problem structure prevalent in large-scale ML tasks. 
To address the challenges posed by variance from zeroth-order gradient estimators and stochastic sampling, we proposed a double variance reduction technique that improves both query efficiency and scalability. 
Our theoretical analysis established that the framework achieves competitive query complexities for both convex and non-convex objectives. 
At the same time, empirical results demonstrated its practical advantages in high-dimensional and large-sample settings. 
The proposed method consistently reduces per-iteration query requirements, offering a practical solution for optimization problems where gradient evaluation is either infeasible or prohibitively expensive. 
This work lays a foundation for further advancements in zeroth-order optimization, particularly in exploring adaptive strategies and extending to more complex problem domains such as distributed or robust optimization.
\vspace{-2mm}


%
%
%

\bibliographystyle{informs2014}

\let\oldbibliography\thebibliography
\renewcommand{\thebibliography}[1]{%
		\oldbibliography{#1}%
		\baselineskip12pt 
		\setlength{\itemsep}{6pt}
	}
\bibliography{reference}

\newpage
\begin{APPENDICES}

\section{Example for the relationship of $L$ and $\hat{L}$}~\label{appendix:example}

\begin{example}\label{Ex:L and HatL}
	Let us consider the following problem:
	\begin{equation}
		f_i(x) = \frac{1}{2}\norm{x}^2 + \frac{n}{2} \cdot a x_i^2, \qquad f(x) = \frac{1}{n}\sum_{i=1}^n f_i(x) =  \frac{1+ a}{2}\norm{x}^2.
	\end{equation} 
	We can observe that $\hL = 1 + na$ and $L = 1 + a$. 
	Thus, it holds that $\frac{\hL}{L} = \frac{1+na }{1+a}$.
\end{example}

\section{Some useful lemmas}

\begin{lemma}[Lemma 1 of \cite{nesterov2017random}]
	Letting the random vector $u \sim \cN(0, I_d)$, for $p\geq 2$, we have
	\begin{equation}\label{eq:u}
		d^{\frac{p}{2}} \leq \EE_u\left[\norm{u}^p\right] \leq (d+p)^{\frac{p}{2}}.
	\end{equation}
\end{lemma}

\begin{lemma}
	Letting $x,y$ be two independent random vectors, then we have
	\begin{equation} \label{eq:xyy}
		\EE\left[\norm{x - y}^2\right] = \EE\left[\norm{x}^2\right] + \EE\left[\norm{y}^2\right], \qquad\mbox{if} \quad \EE[y] = 0.
	\end{equation}
	If $x_i$ with $i = 1, \dots, b$ are of identically independent distribution, then it holds that
	\begin{equation}\label{eq:xx}
		\EE\left[\norm{\frac{1}{b} \sum_{i=1}^{b} x_i - \EE\left[x_i\right]}^2\right] = \frac{1}{b} \cdot \EE\left[\norm{x_i - \EE\left[x_i\right]}^2\right].
	\end{equation}
\end{lemma}

\begin{lemma}
	For two vectors $x, y\in \RR^d$ and given $\alpha > 0$, it holds that
	\begin{align}
		&\dotprod{x, y} \leq \norm{x}\norm{y} \leq \frac{\alpha}{2} \norm{x}^2 + \frac{1}{2\alpha} \norm{y}^2, \label{eq:xy} \\
		&\norm{x+y}^2 \leq (1 + \alpha) \norm{x}^2 + (1+\alpha^{-1}) \norm{y}^2.\label{eq:xy_1}
	\end{align}
\end{lemma}

\begin{lemma}[\cite{magnus1978moments}]\label{lem:prob_res}
	Let $A$ and $B$ be two symmetric matrices, and $u$ has the Gaussian distribution, that is, $u \sim N(0,I_d)$. Define $z = u^\top A u\cdot u^\top B u$. The expectation of $z$ is:
	\begin{equation*}
		\EE_u[z] =  (\tr\;A)(\tr\;B) + 2(\tr\; AB).
	\end{equation*}
\end{lemma}

\begin{lemma}
	Letting $U\in \RR^{d\times \cb}$ be a Gaussian random matrix, $x \in\RR^d$ be a vector and $\alpha>0$ be a scalar, then it holds that
	\begin{equation}\label{eq:UUx}
		\EE\left[\norm{\alpha\cdot UU^\top x - x}^2\right] = \left(1+ \alpha^2 \cb(d+\cb+1) - 2\cb\alpha\right)\cdot \norm{x}^2.
	\end{equation}
\end{lemma}

\textbf{Proof:}
	First, we have
	\begin{align*}
		\EE\left[\norm{\alpha\cdot UU^\top x - x}^2\right] 
		=& \norm{x}^2 + \alpha^2\EE\left[\norm{UU^\top x}^2\right] - 2\alpha\EE\left[x^\top UU^\top x\right] \\
		=& \norm{x}^2 + \alpha^2\EE\left[\norm{UU^\top x}^2\right] - 2\cb\alpha \norm{x}^2.
	\end{align*}
	Letting us denote $u_j = U_{:,j}$, then it holds that $u_j\sim \cN(0, I_d)$.
	Then we have
	\begin{align*}
		\EE\left[\norm{UU^\top x}^2\right]
		=&
		\EE\left[ \norm{\sum_{j=1}^{\cb} u_ju_j^\top x}^2 \right]
		=
		\EE\left[\sum_{i=1}^{\cb}\sum_{j=1}^{\cb} x^\top u_i u_i^\top u_ju_j^\top x \right]\\
		=&
		\EE\left[\sum_{j=1}^{\cb} x^\top u_ju_j^\top x \cdot \norm{u_j}^2 \right] + \sum_{i\neq j}\dotprod{\EE\left[u_iu_i^\top x \right], \EE\left[u_ju_j^\top x\right]}\\
		=&
		\cb (d+2) \norm{x}^2 + \cb(\cb-1)\norm{x}^2\\
		=&
		\cb(d+\cb+1) \norm{x}^2,
	\end{align*}
	where the third equality is because of Lemma~\ref{lem:prob_res}.
	Combining above equations, we can obtain that
	\begin{align*}
		\EE\left[\norm{\alpha\cdot UU^\top x - x}^2\right] = \left(1+ \alpha^2 \cb(d+\cb+1) - 2\cb\alpha\right)\cdot \norm{x}^2.
	\end{align*}
$\hfill\square$

\begin{lemma}[Lemma 3 of \citet{stich2019unified}]
	\label{lem:stich}
	Considering the following sequence
	\begin{align*}
		r_{t+1} \leq (1-\gamma_t) r_t + c\gamma_t^2.
	\end{align*}
	For integer $T$, the stepsize $\gamma_t$, $\gamma_t \leq \frac{1}{a}$, safisfies 
	\begin{align*}
		&\text{if } T \leq a, &\gamma_t &= \frac{1}{a},\\
		&\text{if } T > a  \text{ and } t< t_0, &\gamma_t &= \frac{1}{a},\\
		&\text{if } T > a \text{ and } t \geq t_0, &\gamma_t &= \frac{2}{2a + t - t_0},
	\end{align*}
	where $t_0 = \lceil \frac{T}{2}\rceil$. Then, it holds that
	\begin{align}
		r_{T+1} \leq 32 a r_0\exp(-\frac{T}{2a}) + \frac{36c}{T}.
	\end{align}
\end{lemma}

\begin{lemma}
	Let $f(x)$ be $L$-smooth and the set $\cX$ be convex and bounded with $R^2$. 
	Then the sequence $\{x_t\}$ generated by Algorithm~\ref{alg:zo_fw_vr} satisfies
	\begin{equation} \label{eq:LL}
		\norm{\nabla f(x_t) - \nabla f(x_{t+1})} 
		\leq 
		\gamma_t LR.
	\end{equation}
\end{lemma}
\textbf{Proof:} We have 
\begin{align*}
		\norm{\nabla f(x_t) - \nabla f(x_{t+1})} 
		\leq L \norm{x_t - x_{t+1}}
		= L \gamma_t \norm{s_t - x_t}
		\leq \gamma_t L R.
	\end{align*}
    $\hfill\square$

\section{Proof of Lemma~\ref{lem:nab_h}}

\textbf{Proof of Lemma~\ref{lem:nab_h}:}
By the $L$-smooth assumption, given any $u \sim \cN(0, I_d)$,  we can obtain that 
	\begin{equation}\label{eq:fff}
		\begin{aligned}
			|h(x + \mu u) - \left(h(x) + \mu \dotprod{\nabla h(x), u}\right)  | & \leq \frac{L\mu^2}{2} \norm{u}^2\\
			|h(x - \mu u) - \left(h(x) - \mu_t \dotprod{\nabla h(x), u}\right)  | & \leq  \frac{L\mu^2}{2} \norm{u}^2
		\end{aligned}
	\end{equation}
	Thus, we have
	\begin{equation}\label{eq:nab}
		\begin{aligned}
			\hat{\nabla} h(x, u, \mu) 
			&= \frac{h(x + \mu u) - h(x - \mu u)}{2 \mu}u\\\
			&=  \frac{h(x) + \mu \dotprod{\nabla h(x), u} - \left(h(x) - \mu \dotprod{\nabla h(x), u}  \right)}{2\mu}u + \tau_h(x, \mu, u)\cdot u\\
			&= uu^\top \nabla h(x) +\tau_h(x, u, \mu) \cdot u,
		\end{aligned}
	\end{equation}
	where 
	\begin{align*}
		\tau_h(x, u, \mu) 
		&=  \frac{h(x + \mu u) - h(x-\mu u) -2\mu\dotprod{\nabla h(x), u}}{2\mu}\\
		&=  
		\frac{h(x + \mu u) - \left(h(x) + \mu \dotprod{\nabla h(x), u}\right) - \left(h(x - \mu_t u) - \left(h(x) - \mu \dotprod{\nabla h(x), u}\right)\right) }{2\mu}\\
		&\stackrel{\eqref{eq:fff}}{\leq} \frac{L\mu^2\norm{u}^2}{2\mu}   \\
        &=  \frac{L\mu\norm{u}^2}{2}.
	\end{align*}
	Thus, we have
	\begin{align*}
		\hat{\nabla} h(x, U, \mu) 
		&\stackrel{\eqref{eq:nab_h}}{=}
		\frac{1}{\cb} \sum_{j=1}^\cb \hat{\nabla} h(x, U_{:,j}, \mu)
		\stackrel{\eqref{eq:nab}}{=}
		\frac{1}{\cb}\sum_{j=1}^{\cb} \dotprod{\nabla h(x), U_{:,j}}U_{:,j} + \frac{1}{b} \sum_{j=1}^{b} \tau_h(x, U_{:,j}, \mu)U_{:,j}\\
		&= \frac{1}{b} UU^\top \nabla h(x) + \frac{1}{b} \sum_{j=1}^{b} \tau_h(x, U_{:,j}, \mu)U_{:,j}.
	\end{align*}
$\hfill\square$

\section{Proofs of Section~\ref{subsec:convex}}

\subsection{Proof of Lemma~\ref{lem:f_recursion}}
\textbf{Proof of Lemma~\ref{lem:f_recursion}:}
By the update rule of our algorithm and Assumption~\ref{assump:smoothness}, we have
\begin{align*}
	\quad\quad  & f(x_{t+1}) - f(x^*)\\
	& \leq f(x_t) - f(x^*) + \gamma_t \langle \nabla f(x_t), s_t - x_t\rangle + \frac{L\gamma_t^2}{2}\Vert s_t - x_t\Vert^2\\
	&  = f(x_t) - f(x^*) + \gamma_t \langle \nabla f(x_t) - g_t, s_t - x_t\rangle + \gamma_t \langle g_t, s_t - x_t\rangle + \frac{L\gamma_t^2}{2}\Vert s_t - x_t\Vert^2 \\
	& \leq f(x_t) - f(x^*) + \gamma_t \langle \nabla f(x_t) - g_t, s_t - x_t\rangle + \gamma_t \langle g_t, x^* - x_t\rangle + \frac{L\gamma_t^2}{2}\Vert s_t - x_t\Vert^2\\
	&  =f(x_t) - f(x^*) + \gamma_t \langle \nabla f(x_t) - g_t, s_t - x^* + x^*- x_t\rangle + \gamma_t \langle g_t, x^* - x_t\rangle + \frac{L\gamma_t^2}{2}\Vert s_t - x_t\Vert^2\\
	&  = f(x_t) - f(x^*) + \gamma_t \langle \nabla f(x_t) - g_t, s_t - x^*\rangle + \gamma_t \langle \nabla f(x_t), x^* - x_t\rangle + \frac{L\gamma_t^2}{2}\Vert s_t - x_t\Vert^2\\
	&\leq (1-\gamma_t) (f(x_t)-f(x^*)) + \gamma_t \langle \nabla f(x_t) - g_t, s_t - x^*\rangle + \frac{L\gamma_t^2}{2}\Vert s_t - x_t\Vert^2\\
	&\stackrel{\eqref{eq:xy}}{\leq} (1-\gamma_t) (f(x_t)-f(x^*)) + \frac{\alpha}{2} \Vert g_t - \nabla f(x_t)\Vert^2 + \frac{\gamma_t^2}{2\alpha} \Vert s_t - x^*\Vert^2 + \frac{L\gamma_t^2}{2}\Vert s_t - x_t\Vert^2,
\end{align*}
where the first inequality is because of Assumption~\ref{assump:smoothness}, the second inequality is because of $s_t = \argmin_{s\in \mathcal{S}} \dotprod{s, g_t}$, and  the third inequality is because of Assumption~\ref{assump:convex}.

By Assumption~\ref{assump:bounded set}, we can obtain $\norm{s_t - x^*} \leq R$ and $\norm{s_t - x_t} \leq R$.
Thus, we can obtain that
\begin{equation*}
	f(x_{t+1}) - f(x^*) \leq  (1-\gamma_t) (f(x_t)-f(x^*)) +  \alpha \Vert g_t - \nabla f(x_t)\Vert^2 +\frac{\gamma_t^2R^2}{2\alpha} + \frac{L\gamma_t^2R^2}{2}.
\end{equation*}
$\hfill\square$

\subsection{Proof of Lemma~\ref{lem:ggg}}
\begin{lemma}
	Letting $U\in\RR^{d\times \cb}$ be a random Gaussian matrix and $f(x)$ be a $L$-smooth function, then $\tau_f(x_{t+1}, U_{:,j}, \mu)$ defined in Lemma~\ref{lem:nab_h} satisfies the following property
	\begin{equation}\label{eq:U_up}
		\EE\left[\norm{ \frac{1}{d+\cb+1} \sum_{j=1}^{\cb} \tau_f(x_{t+1}, U_{:,j}, \mu) U_{:,j} }^2\right]
		\leq
		\frac{\cb^2 L^2\mu^2 (d+6)^3}{4(d+\cb+1)^2}.
	\end{equation}
\end{lemma}
\textbf{Proof:}
	By the definition of $\tau_f(\cdot,\cdot,\cdot)$ in Lemma~\ref{lem:nab_h}, we have
	\begin{align*}
		\EE\left[\norm{ \frac{1}{d+\cb+1} \sum_{j=1}^{\cb} \tau_f(x_{t+1}, U_{:,j}, \mu_t) U_{:,j} }^2\right]
		&\leq
		\frac{\cb}{(d+\cb+1)^2} \sum_{j=1}^{\cb} \EE\norm{\tau_f(x_{t+1}, U_{:,j}, \mu) U_{:,j}}^2\\
		&\leq
		\frac{\cb}{(d+\cb+1)^2} \frac{L^2\mu^2}{4} \sum_{j=1}^{\cb} \EE\left[\norm{U_{:,j}}^6\right]\\
		&\stackrel{\eqref{eq:u}}{\leq}
		\frac{\cb^2 L^2\mu^2 (d+6)^3}{4(d+\cb+1)^2}.
	\end{align*}
$\hfill\square$

\begin{lemma}
Letting $U_{t+1}$ be a $d\times \cb$ Gaussian random matrix and $\mu_{t+1} >0$ be a scalar, the approximate gradient $\hat{\nabla} f(x_{t+1}, U_{t+1}, \mu_{t+1})$ is computed as Eq.~\eqref{eq:nab_h}. 
Furthermore, we assume that Assumption~\ref{assump:bounded set}-\ref{assump:smoothness} hold.
Then, we have the following inequality	
\begin{equation}\label{eq:zero_hf_recursion}
\begin{aligned}
	&\EE\left[\norm{	g_t + \frac{\cb}{d+\cb+1} \hat{\nabla} f(x_{t+1}, U_{t+1},\mu_{t+1}) - \frac{U_{t+1}U_{t+1}^\top}{d+\cb+1}g_t - \nabla f(x_{t+1})  }^2\right] \\
	&\leq\left(1-\frac{\cb}{2(d+\cb+1)}\right) \Vert g_t - \nabla f(x_t)\Vert ^2  
	+ \frac{6(d+\cb+1)L^2R^2\gamma_t^2}{\cb} +  \frac{2\cb(d+6)^3}{d+\cb+1} L^2\mu_{t+1}^2.
\end{aligned}
\end{equation}
\end{lemma}
\textbf{Proof:}
	For notation convenience, we use $U$ and $\mu$ instead of $U_{t+1}$ and $\mu_{t+1}$ in our proof.
First, we have
\begin{align*}
	&\EE\left[\norm{g_t + \frac{\cb}{d+\cb+1} \hat{\nabla} f(x_{t+1}, U, \mu) -  \frac{UU^\top}{d+\cb+1} g_{t}-\nabla f(x_{t+1})}^2\right]\\
	&\stackrel{\eqref{eq:naba_hp}}{=}
	\EE\left[ \norm{g_t + \frac{UU^\top}{d+\cb+1} \nabla f(x_{t+1}) -  \frac{UU^\top}{d+\cb+1} g_{t}-\nabla f(x_{t+1})+\frac{1}{d+\cb+1} \sum_{j=1}^{\cb} \tau_f(x_{t+1}, U_{:,j}, \mu) U_{:,j}}^2  \right]\\
	&\stackrel{\eqref{eq:xy_1}}{\leq}
	\left(1 + \frac{\cb}{4(d+\cb+1)}\right) \EE\left[\norm{g_t + \frac{UU^\top}{d+\cb+1} \nabla f(x_{t+1}) -  \frac{UU^\top}{d+\cb+1} g_{t}-\nabla f(x_{t+1})}^2\right] \\
	&+\left(1 + \frac{4(d+\cb +1)}{\cb}\right) \EE\left[\norm{ \frac{1}{d+\cb+1} \sum_{j=1}^{\cb} \tau_f(x_{t+1}, U_{:,j}, \mu) U_{:,j} }^2\right]\\
	&\stackrel{\eqref{eq:UUx}}{=}
	\left(1 + \frac{\cb}{4(d+\cb+1)}\right) \left(1 - \frac{\cb}{d+\cb+1}\right) \norm{g_t - \nabla f(x_{t+1})}^2 \\
    &+\left(1 + \frac{4(d+\cb +1)}{\cb}\right) \EE\left[\norm{ \frac{1}{d+\cb+1} \sum_{j=1}^{\cb} \tau_f(x_{t+1}, U_{:,j}, \mu) U_{:,j} }^2\right]\\
	&\stackrel{\eqref{eq:U_up}}{\leq}
	\left(1 - \frac{3\cb}{4(d+\cb+1)}\right) \norm{g_t - \nabla f(x_{t+1})}^2 + \left(1 + \frac{4(d+\cb +1)}{\cb}\right) \frac{b^2L^2\mu_t^2 (d+6)^3}{4(d+\cb+1)^2}.
\end{align*}

Furthermore, it holds that
\begin{align*}
&\norm{g_t - \nabla f(x_{t+1})}^2 
= \norm{g_t - \nabla f(x_t) +\nabla f(x_t) - \nabla f(x_{t+1})}^2\\
&\stackrel{\eqref{eq:xy_1}}{\leq}
\left(1 + \frac{b}{4(d+b+1)}\right)\norm{g_t - \nabla f(x_t)}^2 + \left(1 + \frac{4(d+b+1)}{b}\right) \norm{\nabla f(x_t) - \nabla f(x_{t+1})}^2\\
&\stackrel{\eqref{eq:LL}}{\le}
\left(1 + \frac{b}{4(d+b+1)}\right)\norm{g_t - \nabla f(x_t)}^2 + \frac{6(d+b+1)L^2R^2\gamma_t^2}{b}. 
\end{align*}
Thus, we can obtain 
\begin{align*}
&\EE\left[\norm{g_t + \frac{\cb}{d+\cb+1} \hat{\nabla} f(x_{t+1}, U, \mu) -  \frac{UU^\top}{d+\cb+1} g_{t}-\nabla f(x_{t+1})}^2\right]\\
&\leq
\left(1 - \frac{3\cb}{4(d+\cb+1)}\right)\left(1 + \frac{b}{4(d+b+1)}\right)\norm{g_t - \nabla f(x_t)}^2 
+ \frac{6(d+b+1)L^2R^2\gamma_t^2}{b}\\ 
&+ \left(1 + \frac{4(d+\cb +1)}{\cb}\right) \frac{b^2L^2\mu_t^2 (d+6)^3}{4(d+\cb+1)^2}\\
&\leq
\left(1-\frac{\cb}{2(d+\cb+1)}\right) \Vert g_t - \nabla f(x_t)\Vert ^2  
+ \frac{6(d+\cb+1)L^2R^2\gamma_t^2}{\cb} +  \frac{2\cb(d+6)^3}{d+\cb+1} L^2\mu^2.
\end{align*}

$\hfill\square$

\begin{lemma}
	Letting $U_{t+1}$ be a $d\times \cb$ Gaussian random matrix and $\mu_{t+1} >0$ be a scalar, the approximate gradients $\hat{\nabla} f_{i_t}(x_{t+1}, U_{t+1}, \mu_{t+1})$ and $\hat{\nabla} f_{i_t}(x_t, U_{t+1}, \mu_{t+1})$ are computed as Eq.~\eqref{eq:nab_h}. 
	We also assume that Assumption~\ref{assump:bounded set}-\ref{ass:hL} hold.
	Then, we have the following inequality
\begin{equation}\label{eq:g_nab}
	\begin{aligned}
		&\EE\left[\norm{ g_t + \frac{1}{|\cS_t|} \sum_{i_t \in \cS_t} \left( \hat{\nabla} f_{i_t}(x_{t+1}, U_{t+1}, \mu_{t+1}) - \hat{\nabla} f_{i_t} (x_t, U_{t+1},\mu_{t+1})\right) - \nabla f(x_{t+1}) }^2\right]\\
		&\leq 
		\left(1 + \frac{p\cb}{4(d+\cb+1)}\right)\cdot\left(\norm{g_t - \nabla f(x_t)}^2 + \left(\frac{(d+\cb+1)\hL^2}{\cb|\cS_t|} + \frac{(d+1)L^2}{\cb}\right)R^2\gamma_t^2\right)\\
		&+\left(1 + \frac{4(d+\cb+1)}{p\cb}\right)\cdot (d+6)^3 \hL^2\mu_{t+1}^2.
	\end{aligned}
\end{equation}
\end{lemma}
\textbf{Proof:}	For notation convenience, we use $U$ and $\mu$ instead of $U_{t+1}$ and $\mu_{t+1}$ in our proof.
	
	First, we have
	\begin{equation}\label{eq:a}
		\begin{aligned}
			&\EE\left[\norm{ g_t + \frac{1}{|\cS_t|} \sum_{i_t \in \cS_t} \left( \hat{\nabla} f_{i_t}(x_{t+1}, U, \mu) - \hat{\nabla} f_{i_t} (x_t, U,\mu)\right) - \nabla f(x_{t+1}) }^2\right]\\
			&\stackrel{\eqref{eq:naba_hp}}{=}
			\EE\left[ \Bigg \lVert g_t +   \frac{UU^\top}{\cb|\cS_t|}\sum_{i_t \in \cS_t} \big(\nabla f_{i_t}(x_{t+1}) - \nabla f_{i_t}(x_t)\big) - \nabla f(x_{t+1}) \right.\\
			&\left.+ \frac{1}{\cb|\cS_t|} \sum_{j=1}^{\cb}\sum_{i_t \in \cS_t} \left(\tau_{f_{i_t}}(x_{t+1}, U_{:,j}, \mu) -  \tau_{f_{i_t}}(x_t, U_{:,j}, \mu)\right)U_{:,j} \Bigg\rVert^2\right]\\
			&\stackrel{\eqref{eq:xy_1}}{\leq}
			\left(1 + \frac{p\cb}{4(d+\cb+1)}\right)\EE\left[ \norm{ g_t + \frac{UU^\top}{\cb|\cS_t|}\sum_{i_t \in \cS_t} \big(\nabla f_{i_t}(x_{t+1}) - \nabla f_{i_t}(x_t)\big) - \nabla f(x_{t+1})  }^2\right]\\
			&+ \left(1 + \frac{4(d+\cb+1)}{p\cb}\right) \EE\left[\norm{\frac{1}{\cb|\cS_t|} \sum_{j=1}^{\cb}\sum_{i_t \in \cS_t} \left(\tau_{f_{i_t}}(x_{t+1}, U_{:,j}, \mu) -  \tau_{f_{i_t}}(x_t, U_{:,j}, \mu)\right)U_{:,j}}^2\right].
		\end{aligned}
	\end{equation}

Next, we will bound the above terms. 
We  have
\begin{align*}
	&\EE\left[ \norm{ g_t + \frac{UU^\top}{\cb|\cS_t|}\sum_{i_t \in \cS_t} \big(\nabla f_{i_t}(x_{t+1}) - \nabla f_{i_t}(x_t)\big) - \nabla f(x_{t+1})  }^2\right]\\
	&=\EE\left[  \norm{ g_t - \nabla f(x_t) + \frac{UU^\top}{\cb|\cS_t|}\sum_{i_t \in \cS_t} \big(\nabla f_{i_t}(x_{t+1}) - \nabla f_{i_t}(x_t)\big) - \big(\nabla f(x_{t+1}) - \nabla f(x_t)\big) }^2 \right]\\
	&\stackrel{\eqref{eq:xyy}}{=}
	\norm{g_t - \nabla f(x_t)}^2 + \EE\left[\norm{ \frac{UU^\top}{\cb|\cS_t|}\sum_{i_t \in \cS_t} \big(\nabla f_{i_t}(x_{t+1}) - \nabla f_{i_t}(x_t)\big) - \big(\nabla f(x_{t+1}) - \nabla f(x_t)\big) }^2\right].
\end{align*}
Furthermore,
\begin{align*}
	&\EE\left[\norm{ \frac{UU^\top}{\cb|\cS_t|}\sum_{i_t \in \cS_t} \big(\nabla f_{i_t}(x_{t+1}) - \nabla f_{i_t}(x_t)\big) - \big(\nabla f(x_{t+1}) - \nabla f(x_t)\big) }^2\right]\\
	&\stackrel{\eqref{eq:xx}}{=}
	\frac{1}{|\cS_t|} \EE\left[ \norm{\frac{UU^\top}{\cb} \big(\nabla f_{i_t}(x_{t+1}) - \nabla f_{i_t}(x_t)\big) - \frac{UU^\top}{\cb}\big(\nabla f(x_{t+1}) - \nabla f(x_t)\big)}^2 \right]
	\\
	& + \EE\left[\norm{ \frac{UU^\top}{\cb}\big(\nabla f(x_{t+1}) - \nabla f(x_t)\big) - \big(\nabla f(x_{t+1}) - \nabla f(x_t)\big)}^2\right]\\
	& \leq
	\frac{1}{|\cS_t|\cb^2}  \cdot  \EE\left[ \norm{ UU^\top \big(\nabla f_{i_t}(x_{t+1}) - \nabla f_{i_t}(x_t)\big)  }^2\right]\\
	&+ \EE\left[\norm{ \frac{UU^\top}{\cb}\big(\nabla f(x_{t+1}) - \nabla f(x_t)\big) - \big(\nabla f(x_{t+1}) - \nabla f(x_t)\big)}^2\right]\\
	&=
	\frac{d+\cb+1}{\cb|\cS_t|} \EE\left[\norm{ \nabla f_{i_t}(x_{t+1}) - \nabla f_{i_t}(x_t) }^2\right] + \frac{d+1}{\cb} \norm{\nabla f(x_{t+1}) - \nabla f(x_t) }^2\\
	&\leq
	\frac{(d+\cb+1)\hL^2}{\cb|\cS_t|} \norm{x_{t+1} - x_t}^2 + \frac{(d+1)L^2}{b} \norm{x_{t+1} - x_t}^2\\
	&=
	\left( \frac{(d+\cb+1)\hL^2}{\cb|\cS_t|} +  \frac{(d+1)L^2}{b}\right) \gamma_t^2 \norm{s_t - x_t}^2,
\end{align*}
where the last inequality is because of Assumption~\ref{ass:hL} and Assumption~\ref{assump:smoothness}.
Combining with Assumption~\ref{assump:bounded set}, we can obtain that
\begin{align*}
	&\EE\left[  \norm{ \frac{UU^\top}{\cb|\cS_t|}\sum_{i_t \in \cS_t} \big(\nabla f_{i_t}(x_{t+1}) - \nabla f_{i_t}(x_t)\big) - \big(\nabla f(x_{t+1}) - \nabla f(x_t)\big) }^2 \right]\\
	&\leq
	\left( \frac{(d+\cb+1)\hL^2}{\cb|\cS_t|} +  \frac{(d+1)L^2}{b}\right) \gamma_t^2R^2.
\end{align*}
Therefore, we can obtain that
\begin{equation}\label{eq:aa}
	\begin{aligned}
		&\EE\left[ \norm{ g_t + \frac{UU^\top}{\cb|\cS_t|}\sum_{i_t \in \cS_t} \big(\nabla f_{i_t}(x_{t+1}) - \nabla f_{i_t}(x_t)\big) - \nabla f(x_{t+1})  }^2\right]\\
		&\leq 
		\norm{g_t - \nabla f(x_t)}^2 + \left( \frac{(d+\cb+1)\hL^2}{\cb|\cS_t|} +  \frac{(d+1)L^2}{b}\right) \gamma_t^2R^2.
	\end{aligned}
\end{equation}

Next, we can bound that 
\begin{equation}\label{eq:aaa}
\begin{aligned}
	&\EE\left[\norm{\frac{1}{\cb|\cS_t|} \sum_{j=1}^{\cb}\sum_{i_t \in \cS_t} \left(\tau_{f_{i_t}}(x_{t+1}, U_{:,j}, \mu) -  \tau_{f_{i_t}}(x_t, U_{:,j}, \mu)\right)U_{:,j}}^2\right]\\
	&\leq
	2 \EE\left[ \norm{ \frac{1}{\cb|\cS_t|} \sum_{j=1}^{\cb}\sum_{i_t \in \cS_t} \tau_{f_{i_t}}(x_{t+1}, U_{:,j}, \mu) U_{:,j} }^2 + \norm{ \frac{1}{\cb|\cS_t|} \sum_{j=1}^{\cb}\sum_{i_t \in \cS_t}  \tau_{f_{i_t}}(x_t, U_{:,j}, \mu) U_{:,j} }^2 \right]\\
	&\leq
	\frac{2}{\cb|\cS_t|}\sum_{j=1}^{\cb} \sum_{i_t \in \cS_t} \EE\left[ \norm{\tau_{f_{i_t}}(x_{t+1}, U_{:,j}, \mu) U_{:,j}}^2 + \norm{ \tau_{f_{i_t}}(x_t, U_{:,j}, \mu) U_{:,j}}^2 \right]\\
	&\leq
	\frac{2}{\cb|\cS_t|}\sum_{j=1}^{\cb} \sum_{i_t \in \cS_t}  \frac{\hL^2\mu^2}{2} \EE\left[\norm{U_{:,j}}^6\right]
	\stackrel{\eqref{eq:u}}{\leq} (d+6)^3\hL^2\mu^2,
\end{aligned}
\end{equation}
where the last inequality is because of Lemma~\ref{lem:nab_h} with $f_i(x)$ being $\hL$-smooth.

Combining Eq.~\eqref{eq:a}, \eqref{eq:aa}, Eq.~\eqref{eq:aaa}, we can obtain that
\begin{align*}
&\EE\left[\norm{ g_t + \frac{1}{|\cS_t|} \sum_{i_t \in \cS_t} \left( \hat{\nabla} f_{i_t}(x_{t+1}, u_t, \mu_t) - \hat{\nabla} f_{i_t} (x_t, u_t,\mu)\right) - \nabla f(x_{t+1}) }^2\right]\\
&\stackrel{\eqref{eq:a}}{\leq}
\left(1 + \frac{p\cb}{4(d+\cb+1)}\right)\EE\left[ \norm{ g_t + \frac{UU^\top}{\cb|\cS_t|}\sum_{i_t \in \cS_t} \big(\nabla f_{i_t}(x_{t+1}) - \nabla f_{i_t}(x_t)\big) - \nabla f(x_{t+1})  }^2\right]\\
&+ \left(1 + \frac{4(d+\cb+1)}{p\cb}\right) \EE\left[\norm{\frac{1}{\cb|\cS_t|} \sum_{j=1}^{\cb}\sum_{i_t \in \cS_t} \left(\tau_{f_{i_t}}(x_{t+1}, U_{:,j}, \mu) -  \tau_{f_{i_t}}(x_t, U_{:,j}, \mu)\right)U_{:,j}}^2\right]\\
&\stackrel{\eqref{eq:aa}\eqref{eq:aaa}}{\leq}
\left(1 + \frac{p\cb}{4(d+\cb+1)}\right) \left(\norm{g_t - \nabla f(x_t)}^2 + \left( \frac{(d+\cb+1)\hL^2}{\cb|\cS_t|} +  \frac{(d+1)L^2}{b}\right) \gamma_t^2R^2\right)\\
&+\left(1 + \frac{4(d+\cb+1)}{p\cb}\right)(d+6)^3\hL^2\mu^2.
\end{align*}
$\hfill\square$

Based on the above two lemmas, we will provide detailed proof of Lemma~\ref{lem:ggg}.

\textbf{Proof of Lemma~\ref{lem:ggg}:} By the update of $g_t$, we can obtain that
	\begin{align*}
		&\EE\left[\norm{g_{t+1} - \nabla f(x_{t+1})}^2\right]\\
		&=
		p \cdot \EE\left[\norm{	g_t + \frac{\cb}{d+\cb+1} \hat{\nabla} f(x_{t+1}, U_{t+1},\mu_{t+1}) - \frac{U_{t+1}U_{t+1}^\top}{d+\cb+1}g_t - \nabla f(x_{t+1})  }^2\right]\\
		&+(1 - p)\cdot \EE\left[\norm{ g_t + \frac{1}{|\cS_t|} \sum_{i_t \in \cS_t} \left( \hat{\nabla} f_{i_t}(x_{t+1}, U_{t+1}, \mu_{t+1}) - \hat{\nabla} f_{i_t} (x_t, U_{t+1},\mu_{t+1})\right) - \nabla f(x_{t+1}) }^2\right].
	\end{align*}
	
	Thus, we can obtain that
	\begin{align*}
		&\EE\left[\norm{g_{t+1} - \nabla f(x_{t+1})}^2\right] \\
		&\leq
		p\left(1-\frac{\cb}{2(d+\cb+1)}\right) \Vert g_t - \nabla f(x_t)\Vert ^2  
		+
		\frac{6p(d+\cb+1)L^2R^2\gamma_t^2}{\cb} +  \frac{2p\cb(d+6)^3}{d+\cb+1} L^2\mu_{t+1}^2\\
		&+(1 - p)\left(1 + \frac{p\cb}{4(d+\cb+1)}\right)\cdot\left(\norm{g_t - \nabla f(x_t)}^2 + \left(\frac{(d+\cb+1)\hL^2}{\cb|\cS_t|} + \frac{(d+1)L^2}{\cb}\right)R^2\gamma_t^2\right)\\
		&+(1 - p)\left(1 + \frac{4(d+\cb+1)}{p\cb}\right)\cdot (d+6)^3 \hL^2\mu_{t+1}^2\\
		&\leq
		\left(1 - \frac{p\cb}{4(d+\cb+1)}\right) \norm{g_t - \nabla f(x_t)}^2 + \left(\frac{6p(d+\cb+1)L^2}{\cb} + \frac{2(d+\cb+1)\hL^2}{\cb|\cS_t|} + \frac{2(d+1)L^2}{\cb}\right) R^2\gamma_t^2\\
		&+\max\left\{\frac{8(d+\cb+1)(d+6)^3}{p\cb}\hL^2,\; \frac{2p\cb(d+6)^3}{d+\cb+1}L^2
		\right\} \cdot \mu_{t+1}^2\\
		&\leq
		\left(1 - \frac{p\cb}{4(d+\cb+1)}\right) \norm{g_t - \nabla f(x_t)}^2 + \left(\frac{6p(d+\cb+1)L^2}{\cb} + \frac{2(d+\cb+1)\hL^2}{\cb|\cS_t|} + \frac{2(d+1)L^2}{\cb}\right) R^2\gamma_t^2\\
		&+\frac{8(d+\cb+1)(d+6)^3}{p\cb}\hL^2\mu_{t+1}^2.
	\end{align*}
$\hfill\square$

\section{Proofs of Section~\ref{subsec:nonconv}}

\subsection{Proof of Lemma~\ref{lem:nonconv}}

\textbf{Proof of Lemma~\ref{lem:nonconv}:}
	First, by the $L$-smoothness of the objective function in Assumption~\ref{assump:smoothness}, we can obtain that,
	\begin{align*}
		f(x_{t+1}) \leq& f(x_t) + \langle \nabla f(x_t), x_{t+1} - x_t\rangle + \frac{L}{2}\Vert x_{t+1} - x_t\Vert^2\\
		=& f(x_t) + \gamma_t \langle \nabla f(x_t), s_t - x_t\rangle + \frac{L\gamma_t^2}{2}\Vert s_t - x_t\Vert^2\\
		=& f(x_t) + \gamma_t \langle \nabla f(x_t) - g_t, s_t - x_t\rangle + \gamma_t \langle g_t, s_t - x_t\rangle + \frac{L\gamma_t^2}{2}\Vert s_t - x_t\Vert^2\\
		\leq& f(x_t) + \gamma_t \langle \nabla f(x_t) - g_t, s_t - x_t\rangle + \gamma_t \langle g_t, s_t - x_t\rangle + \frac{L\gamma_t^2R^2}{2},
	\end{align*}
	where the last inequality is because of Assumption~\ref{assump:bounded set}.
	
	The optimal choice of $s_t$ in Algorithm~\ref{alg:zo_fw_vr} gives that $\dotprod{g_t, s_t - x_t} \leq \dotprod{g_t, x-x_t}$ for all $x\in \cX$. 
	Then, we have
	\begin{align*}
		&f(x_{t+1}) - f(x^*) \\
		&\leq
		f(x_t) - f(x^*)  + \gamma_t \langle \nabla f(x_t) - g_t, s_t - x_t\rangle + \gamma_t \langle g_t, s_t - x_t\rangle + \frac{L\gamma_t^2R^2}{2}\\
		&\leq
		f(x_t) - f(x^*)  + \gamma_t \langle \nabla f(x_t) - g_t, s_t - x_t\rangle + \gamma_t \langle g_t, x - x_t\rangle + \frac{LR^2\gamma_t^2}{2}\\
		&=
		f(x_t) - f(x^*)  + \gamma_t \dotprod{\nabla f(x_t) - g_t, s_t - x_t} +\gamma_t \dotprod{g_t -\nabla f(x_t), x-x_t}\\
		&+\gamma_t\dotprod{\nabla f(x_t), x- x_t} + \frac{L\gamma_t^2R^2}{2}\\
		&=
		f(x_t) - f(x^*)  + \gamma_t\dotprod{\nabla f(x_t) - g_t, s_t - x} + \gamma_t \dotprod{\nabla f(x_t), x - x_t} + \frac{LR^2\gamma_t^2}{2}\\
		&\stackrel{\eqref{eq:xy}}{\leq}
		f(x_t) - f(x^*)  + \gamma_t\dotprod{\nabla f(x_t), x-x_t} + \gamma_t \norm{\nabla f(x_t) - g_t} \cdot \norm{s_t - x} + \frac{LR^2\gamma_t^2}{2} \\
		&\leq f(x_t) - f(x^*)  + \gamma_t\dotprod{\nabla f(x_t), x-x_t} + \gamma_t R\cdot \norm{\nabla f(x_t) - g_t}  + \frac{LR^2\gamma_t^2}{2},
	\end{align*}
	where the last inequality is because of Assumption~\ref{assump:bounded set}. Thus, we can obtain 
	\begin{equation*}
		\gamma_t\dotprod{\nabla f(x_t), x_t-x} 
		\leq f(x_t) - f(x^*) - \Big(f(x_{t+1}) - f(x^*)  \Big) + \gamma_t R\cdot \norm{\nabla f(x_t) - g_t}  + \frac{LR^2\gamma_t^2}{2},
	\end{equation*}
	which proves Eq.~\eqref{eq:ggap}.
	
	Using Eq.~\eqref{eq:xy} to the above equation, we can obtain that
	\begin{align*}
		\gamma_t\dotprod{\nabla f(x_t), x-x_t} 
		\leq 
		f(x_t) - f(x^*) - \big(f(x_{t+1}) - f(x^*)\big) + \frac{\alpha}{L}\norm{\nabla f(x_t) - g_t}^2 + \frac{LR^2\gamma_t^2}{\alpha}  + \frac{LR^2\gamma_t^2}{2}.
	\end{align*}
$\hfill\square$

\subsection{Proof of Lemma~\ref{lem:var}}

\textbf{Proof of Lemma~\ref{lem:var}:}
	By choosing $\gamma_t = \frac{1}{\sqrt{T}}$ and $\mu_{t+1} = \sqrt{\frac{p}{|\cS|(d+6)^3 T}}R$, Eq.~\eqref{eq:ggg} reduces to 
	\begin{align*}
		&\EE\left[\norm{g_{t+1} - \nabla f(x_{t+1})}^2\right] \\
		&\leq
		\left(1 - \frac{p\cb}{4(d+\cb+1)}\right) \EE\left[\norm{g_t - \nabla f(x_t)}^2\right] + \left(\frac{6p(d+\cb+1)L^2}{\cb} + \frac{2(d+1)L^2}{\cb} + \frac{10(d+\cb+1)\hL^2}{\cb|\cS|}\right) \frac{R^2}{T}\\
		&\leq
		\left(1 - \frac{p\cb}{4(d+\cb+1)}\right)^{t+1} \norm{g_0 - \nabla f(x_0)}^2 \\
		&+\left(\frac{6p(d+\cb+1)L^2}{\cb} + \frac{2(d+1)L^2}{\cb} + \frac{10(d+\cb+1)\hL^2}{\cb|\cS|}\right) \frac{R^2}{T}\sum_{i=0}^{t-1} \left(1 - \frac{p\cb}{4(d+\cb+1)}\right)^i\\
		&\leq
		\left(1 - \frac{p\cb}{4(d+\cb+1)}\right)^{t+1} \norm{g_0 - \nabla f(x_0)}^2 \\
		&+ \left(\frac{6p(d+\cb+1)L^2}{\cb} + \frac{2(d+1)L^2}{\cb} + \frac{10(d+\cb+1)\hL^2}{\cb|\cS|}\right) \frac{R^2}{T} \cdot \frac{4(d+\cb+1)}{p\cb} \\
		&\leq
		\left(1 - \frac{p\cb}{4(d+\cb+1)}\right)^{t+1} \norm{g_0 - \nabla f(x_0)}^2 + \left(\frac{32(d+\cb+1)^2L^2}{p\cb^2} + \frac{40(d+\cb+1)^2\hL^2}{p|\cS|\cb^2} \right) \cdot \frac{R^2}{T}.
	\end{align*}
$\hfill\square$

\section{Experiment Setting}\label{appendix:setting}

\setlength{\tabcolsep}{7pt} 
\renewcommand{\arraystretch}{1.1} 

\begin{table}[!htb]
    \caption{Characteristics of Datasets}
    \label{tab:dataset_info_r}
    \centering
    \footnotesize
    \begin{tabular}{lcccc}
        \hline
        Dataset & $n$ & $d$ & \# Classes & Type \\ 
        \hline
        RCV1        & 20,242 & 47,236 & 2  & Text \\
        Real-Sim    & 72,309 & 20,958 & 2  & Text \\
        a9a         & 32,561 & 123    & 2  & Tabular \\
        w8a         & 49,749 & 300    & 2  & Tabular \\
        MNIST (Class 1) & 1,131  & 784    & 1 & Image \\
        CIFAR       & 45,628 & 3,072  & 10 & Image \\
        \hline
    \end{tabular}
    \vspace{-0.1in}
\end{table}

In this paper, we evaluate our optimization algorithm with six widely adopted benchmark datasets: RCV1, Real-Sim, a9a, w8a, MNIST, and CIFAR-10 datasets. In the sparse logistic regression and robust classification experiments, we use four datasets obtained from LIBSVM website\footnote{\url{https://www.csie.ntu.edu.tw/~cjlin/libsvmtools/datasets/binary.html}}. The RCV1 and Real-sim datasets are text datasets, where features represent textual information, and labels correspond to binary document categories. The a9a (Adult) dataset consists of tabular demographic features, such as age, education level, and occupation, with a binary classification label indicating whether an individual's annual income exceeds \$50K. Similarly, the w8a dataset contains several web page features with binary classification labels. 

In the black-box attack experiments, we evaluate the performance on MNIST~\citep{lecun2010mnist} and CIFAR-10~\citep{krizhevsky2009cifar} image datasets.  For the MNIST dataset, we employ a well-trained DNN model\footnote{\url{https://github.com/carlini/nn_robust_attacks}} as the target classifier.  For the CIFAR dataset, we pre-train a ResNet-18 classifier\footnote{\url{https://github.com/YingHuiH/ZSFW-DVR}}. The statistics of these datasets are summarized in Table \ref{tab:dataset_info_r}.



For all tasks, we maintain consistency in the batch size and the number of directions $b$ for our algorithm and ZOFWSGD, as detailed in Table \ref{tab:setting_param}. Similarly, we use the same batch size and the corresponding epoch size $q$ for AccSZOFW. To increase the number of iterations, we reduce the number of batch sizes and the number of directions $b$ in the robust black-box classification and adversarial attack (nonconvex objective function) compared to the black-box sparse logistic regression (convex objective function).
For the black-box sparse logistic regression and black-box robust classification, our algorithm ZOFW-DVR adopt a learning rate schedule of $\mathcal{O}(lr/(t+1))$, following \citet{sahu2019towards}. For the black-box constrained adversarial attack, we use a fixed learning rate.   
Each method's learning rate $lr$ is individually fine-tuned for each dataset to ensure optimal performance.




Finally, our experiments are conducted on a server equipped with an 18 vCPU AMD EPYC 9754 128-Core Processor and an RTX 4090D GPU , as well as a Mac Studio with an M2 Max chip.

\begin{table}[!htb]
    \caption{Parameter Settings for Different Experiments}
    \label{tab:setting_param}
    \centering
    \footnotesize
    \begin{tabular}{lcccccc}
        \hline
        Experiment & Dataset & $r$ & Constraint & Batch Size & $b$ & $q$ \\
        \hline
        \multirow{4}{*}{Black-box Sparse Logistic Regression} 
        & RCV1       & 20 & $||x||_1 < r$ & 200 & 400 & 142 \\
        & Real-Sim   & 20 & $||x||_1 < r$ & 400 & 200 & 268 \\
        & a9a        & 2  & $||x||_1 < r$ & 200 & 20  & 180 \\
        & w8a        & 2  & $||x||_1 < r$ & 200 & 30  & 223 \\
        \hline
        \multirow{4}{*}{Black-box Robust Classification} 
        & RCV1       & 20 & $||x||_1 < r$ & 143 & 218 & 143 \\
        & Real-Sim   & 20 & $||x||_1 < r$ & 268 & 144 & 268 \\
        & a9a        & 2  & $||x||_1 < r$ & 180 & 12  & 175 \\
        & w8a        & 2  & $||x||_1 < r$ & 223 & 18  & 223 \\
        \hline
        \multirow{2}{*}{Black-box Constrained Adversarial Attack} 
        & MNIST (Class 1) & 10 & $||x||_2 < r$ & 33  & 28  & 33 \\
        & CIFAR      & 5  & $||x||_2 < r$ & 70  & 55  & 70 \\
        \hline
    \end{tabular}
    \vspace{-0.1in}
\end{table}

\begin{figure}[htp]
\centering

\subfloat[MNIST (Class 1)]{%
  \includegraphics[width=0.48\textwidth]{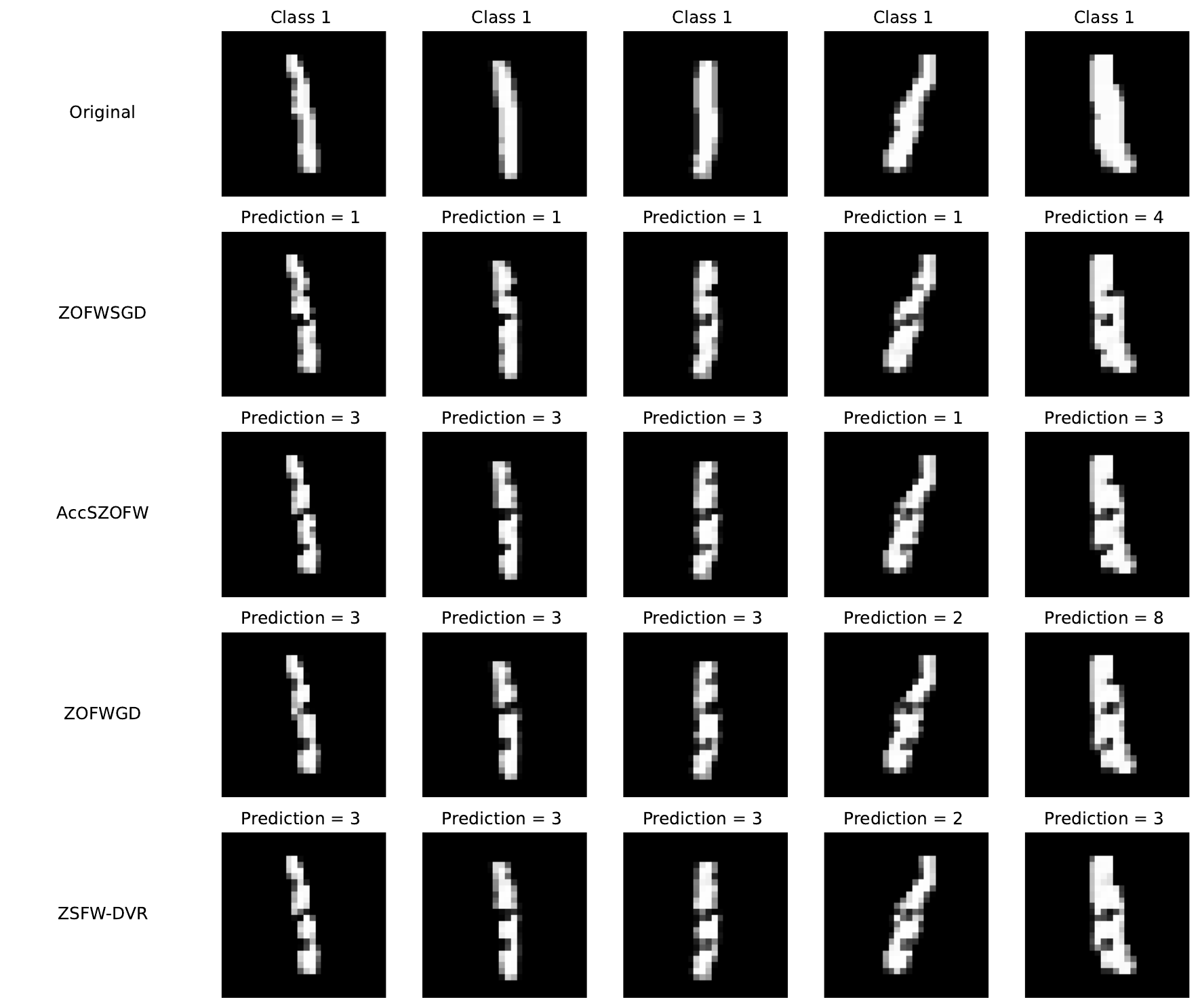}%
}
\subfloat[CIFAR-10]{%
  \includegraphics[width=0.48\textwidth]{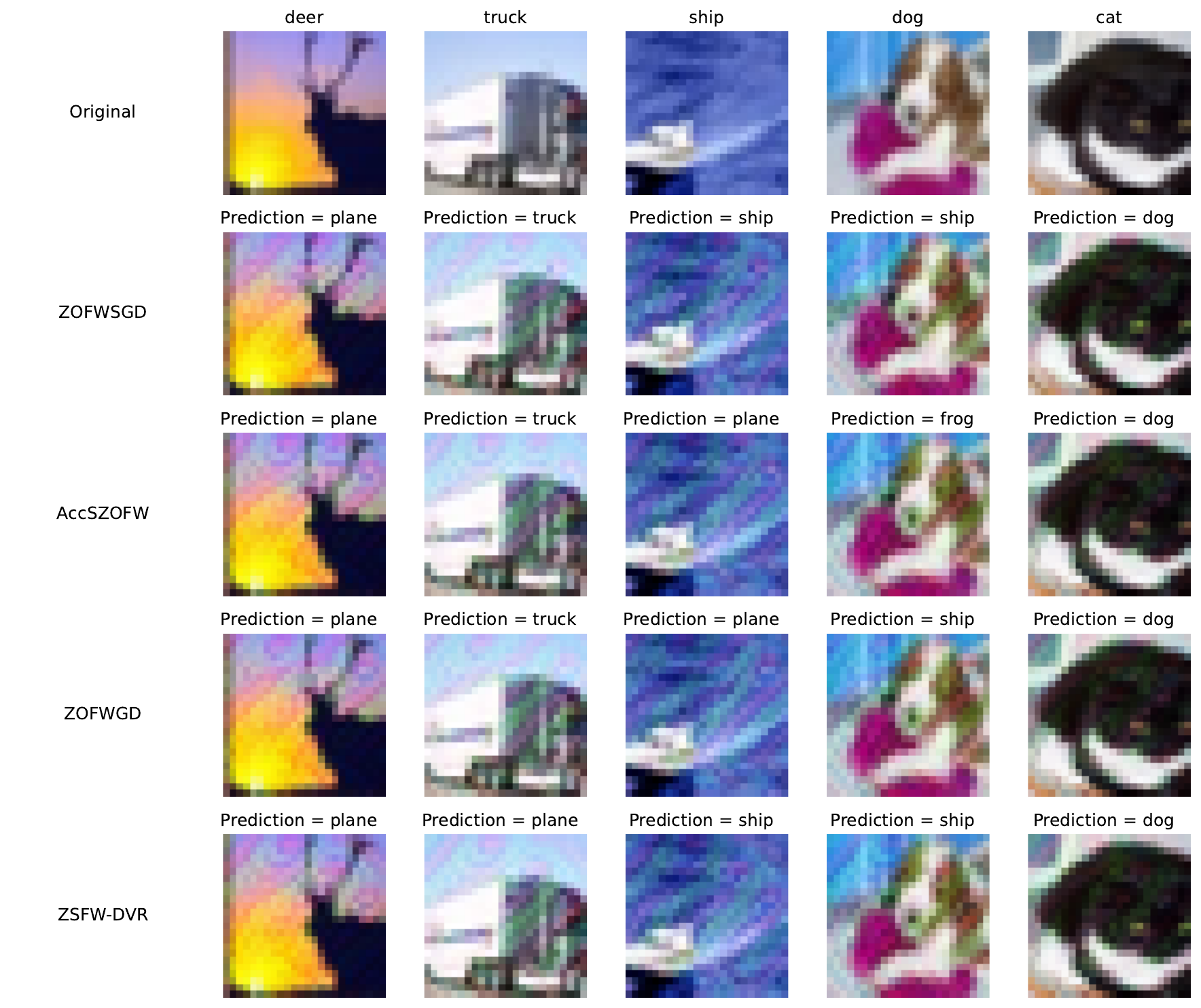}%
}

\caption{Visual comparisons of original images and their adversarial examples generated by each algorithm. }
\label{fig:img_visual}

\end{figure}

\end{APPENDICES}



\end{document}